\documentclass[letterpaper, 10 pt, conference]{ieeeconf}    % Use this line for a4 paper
\IEEEoverridecommandlockouts % This command is only needed if % you want to use the \thanks command
\usepackage[utf8]{inputenc}
\usepackage{amsmath}
\usepackage{amssymb}
\usepackage[position=bottom]{subfig}
\usepackage[font=small]{caption}
\usepackage{amsfonts}
\usepackage{mathtools}
\usepackage{url}
\usepackage{bm}
\usepackage{enumerate}
\usepackage{xcolor}
\usepackage{graphicx}
\usepackage{color}

\title{2022-deformable-elastic-objects-UAVs}
\date{Jan 2025}

\title{\LARGE \bf Manipulation of Elasto-Flexible Cables with Single or Multiple UAVs }
\author{Chiara Gabellieri$^{1}$, Lars Teeuwen$^1$, Yaolei Shen$^1$, Antonio Franchi$^{1,2}$
\thanks{$^1$ Robotics and Mechatronics Department, Electrical Engineering,  Mathematics, and Computer Science (EEMCS) Faculty, University of Twente, 7500 AE Enschede, The Netherlands. {\tt\small c.gabellieri@utwente.nl}, {\tt\footnotesize a.franchi@utwente.nl}}
\thanks{$^2$ Department of Computer, Control and Management Engineering, Sapienza University of Rome, 00185 Rome, Italy, {\tt\footnotesize antonio.franchi@uniroma1.it}}
\thanks{This work was partially funded by the Horizon Europe research and innovation programs under agreement no. 101059875 (FLYFLIC),  agreement no. 101120732 (AUTOASSESS), and agreement no. 101182891 (NEUTRAWEED).}}
\begin{document} 

% ---------- Standard Commands ------------------

\newcommand\red[1]{{\textcolor{red}{#1}}}
\newcommand\blue[1]{{\textcolor{blue}{#1}}}
\newcommand\green[1]{{\textcolor{green}{#1}}}
% ---------- new variables ------------------

% Sensitivity
\newcommand{\errorRL}{e_{\rotMatL}}
% ---------- new commands ------------------
\newcommand{\vect}[1]{\bm{#1}}		% vectors
\newcommand{\matr}[1]{\bm{#1}}		% matrices
\newcommand{\nR}[1]{\mathbb{R}^{#1}}		% real number
\newcommand{\nN}[1]{\mathbb{N}^{#1}}		% inter number
\newcommand{\SO}[1]{SO({#1})}		% orthogonal group
\newcommand{\define}{:=}			% define symbol
\newcommand{\modulus}[1]{\left| #1 \right|}	% abs
\newcommand{\matrice}[1]{\begin{bmatrix} #1 \end{bmatrix}}	% matrix
\newcommand{\smallmatrice}[1]{\left[\begin{smallmatrix} #1 \end{smallmatrix}\right]}	% matrix
\newcommand{\cosp}[1]{\cos \left( #1 \right)}	% cos with brace
\newcommand{\sinp}[1]{\sin \left( #1 \right)}	% sin with brace
\newcommand{\determinant}[1]{\text{det}\left(#1\right)} 	% determinant
\newcommand{\sgn}[1]{\text{sgn}\left( #1 \right)}			% signum
\newcommand{\atanTwo}[1]{{\rm atan2}\left( #1\right)}		% atan2
\newcommand{\acotTwo}[1]{{\rm acot2}\left( #1\right)}		% acot2
\newcommand{\upperRomannumeral}[1]{\uppercase\expandafter{\romannumeral#1}}	% roman numbers
\newcommand{\lowerromannumeral}[1]{\romannumeral#1\relax}
\newcommand{\vSpace}{\;\,}
\newcommand{\image}[1]{\text{Im}\left( #1 \right)}
\newcommand{\pinv}{\dagger}
\newcommand{\diag}[1]{\text{diag}\left( #1 \right)}
\newcommand{\unitOfMeasure}[1]{\; [ \rm #1]}
\newcommand{\Ker}{{\rm Null}}
\newcommand{\ith}[1]{#1^{\rm{th}}}
\newcommand{\jth}{j^{\rm{th}}}
% --------- References ----------------------
\newcommand{\fig}{Fig.~}	% figure ref
\newcommand{\eqn}{Eq.~}	% equation ref
\newcommand{\tab}{Tab.~}	% table ref
\newcommand{\cha}{Chap.~}	% chapter ref
\newcommand{\sect}{Sec.~}	% section ref
\newcommand{\theo}{Result~}	% section ref
\newcounter{example}[section]
\newenvironment{example}[1][]{\refstepcounter{example}\par\medskip
   \noindent \textbf{Result~\theexample. #1} \rmfamily}{\medskip}
% --------- Variables -----------------------

% General
\renewcommand{\frame}{\mathcal{F}}		% frame
\newcommand{\origin}{O}						% origin
\newcommand{\vX}{\vect{x}}					% x-axis
\newcommand{\vY}{\vect{y}}					% y-axis
\newcommand{\vZ}{\vect{z}}					% z-axis
\newcommand{\vE}[1]{\vect{e}_{#1}}			% standard unit axis
\newcommand{\vV}{\vect{v}}					% general vector
\newcommand{\pos}{\vect{p}}				% position vector
\newcommand{\dpos}{\dot{\vect{p}}}		% derivative of position vector
\newcommand{\ddpos}{\ddot{\vect{p}}}	% double

\newcommand{\err}{\vect{e}}				% position vector
\newcommand{\derr}{\dot{\vect{e}}}		% derivative of position vector
\newcommand{\dderr}{\ddot{\vect{e}}}%derivative of position vector
\newcommand{\rotMat}{\matr{R}}			% rotation matrix
\newcommand{\rotMatVectAngle}[2]{\rotMat_{#1}(#2)}	% rotation matrix representing the rotation about a vector of a certain angle
\newcommand{\angVel}{\vect{\omega}}				% angular velocity
\newcommand{\angAcc}{\dot{\vect{\omega}}}		% angular acceleration
\newcommand{\vZero}{\vect{0}}				% vect/matr of zeros
\newcommand{\gravity}{\vect{g}}			% gravity vector
\renewcommand{\skew}[1]{\matr{S}(#1)}				% sckew operator
\newcommand{\eye}[1]{\matr{I}_{#1}}		% identity matrix
\newcommand{\roll}{\phi}		% roll
\newcommand{\pitch}{\theta}		% pitch
\newcommand{\yaw}{\psi}		% yaw
\newcommand{\eulerAngles}{\vect{\eta}}

% World frame
\newcommand{\frameW}{\frame_W}			% world frame
\newcommand{\originW}{\origin_W}		% origin world frame
\newcommand{\xW}{\vX_W}				% x-axis world frame
\newcommand{\yW}{\vY_W}				% y-axis world frame
\newcommand{\zW}{\vZ_W}				% z-axis world frame

% Load
\newcommand{\frameL}{\frame_L}			% load frame
\newcommand{\originL}{\origin_L}			% origin load frame
\newcommand{\xL}{\vX_L}				% x-axis load frame
\newcommand{\yL}{\vY_L}				% y-axis load frame
\newcommand{\zL}{\vZ_L}				% z-axis load frame
\newcommand{\pL}{\pos_L}			% load position
\newcommand{\pLW}{\prescript{W}{}{\pL}} % load position wrt world frame (explicit)
\newcommand{\dpL}{\dpos_L}			% load  velocity position
\newcommand{\ddpL}{\ddpos_L}		% load acceleration position
\newcommand{\rotMatL}{\rotMat_L}% rotation matrix for load orientation
\newcommand{\rotMatLEquilib}{\rotMat^{eq}_{L}}
\newcommand{\rotMatLInc}{\hat{\rotMat}_L}
\newcommand{\rotMatLW}{\prescript{W}{}{\rotMat_L}}	% rotation matrix for load orientation wrt world frame (explicit)
\newcommand{\angVelL}{\angVel_L}		% angular velocity
\newcommand{\angAccL}{\angAcc_L}		% angular acceleration
\newcommand{\massL}{{m_L}}
\newcommand{\massLU}{\hat{{m}}_L}				% load mass
\newcommand{\inertiaL}[1]{\matr{J}_{#1}}	% load inertia
\newcommand{\InertiaL}{\matr{M}_L}	% general load inertia
\newcommand{\coriolisL}{\vect{c}_L}	% coriolis terms of the load
\newcommand{\gravityL}{\vect{g}_L}	% gravity terms of the load	
\newcommand{\graspL}{\matr{G}}		% grasp matrix of the load	
\newcommand{\dampingL}{\matr{B}_L}	% damping coeff
\newcommand{\deriv}[1]{^{(#1)}}
% Cables
\newcommand{\length}[1]{{l}_{0#1}}
\newcommand{\lengthU}[1]{\hat{{l}}_{0#1}}	% nominal length of the cable
\newcommand{\springCoeff}[1]{{k}_{#1}}

\newcommand{\springCoeffbar}[1]{\bar{k}_{#1}}
\newcommand{\springCoeffU}[1]{\hat{{k}}_{#1}}% spring coefficent
\newcommand{\cableForce}[1]{\vect{f}_{#1}}
\newcommand{\dCableForce}[1]{\dot{\vect{f}}_{#1}}
\newcommand{\cableForceEquilib}[1]{\vect{f}^{eq}_{#1}}
\newcommand{\cableForceInc}[1]{\hat{\vect{f}}_{#1}}			% cable force
\newcommand{\cableForceU}[1]{\hat{\vect{f}}_{#1}}
\newcommand{\cableAttitudeNorm}[1]{\vect{n}_{#1}}% cable attitude
\newcommand{\cableAttitude}[1]{\vect{l}_{#1}}% cable attitude
\newcommand{\dcableAttitude}[1]{\dot{\vect{l}}_{#1}}	% dot cable attitude

\newcommand{\condZero}{\xi}
\newcommand{\anchorPoint}[1]{B_{#1}}			% attachement point to the load
\newcommand{\anchorPos}[1]{\vect{b}_{#1}}		% attachement position to the load
\newcommand{\anchorLength}[1]{{b}_{#1}}	
\newcommand{\anchorLengthU}[1]{\hat{{b}}_{#1}}		% attachement position to the load
\newcommand{\anchorPosL}[1]{\prescript{L}{}{\vect{b}}_{#1}}		% attachement position to the load wrt P
\newcommand{\anchorPosLU}[1]{\prescript{L}{}{\hat{\vect{b}}}_{#1}}		% attachement position to the load wrt P

\newcommand{\robotPoint}[1]{A_{#1}}				% attachment point to the robot
\newcommand{\robotPos}[1]{\vect{a}_{ #1}}		% attachement position to the robot
\newcommand{\robotPosP}[1]{\prescript{P}{}{\vect{a}}_{#1}}		% attachement position to the load wrt P
\newcommand{\angleCable}[1]{\alpha_{#1}}		% generalized coordinate of the cable
\newcommand{\angleCables}{\vect{\alpha}}		% vector of all the angles of the cables
\newcommand{\tension}[1]{t_{#1}}				% tension
\newcommand{\tensionMax}[1]{\overline{f}_{L#1}}				% tension max
\newcommand{\tensionMin}[1]{\underline{f}_{L#1}}				% tension min
\newcommand{\cableForces}{\cableForce{}}		% cable forces

% Robot
\newcommand{\frameR}[1]{\frame_{R #1}}			% robot frame
\newcommand{\originR}[1]{A_{#1}}					% origin robot frame
\newcommand{\xR}[1]{\vX_{R #1}}								% x-axis robot frame
\newcommand{\yR}[1]{\vY_{R #1}}								% y-axis robot frame
\newcommand{\zR}[1]{\vZ_{R #1}}								% z-axis robot frame
\newcommand{\pR}[1]{\pos_{R #1}}						% robot position
\newcommand{\dpR}[1]{\dpos_{R #1}}					% robot velocity
\newcommand{\ddpR}[1]{\ddpos_{R #1}}				% robot acceleration
\newcommand{\uR}[1]{\vect{u}_{R #1}}				% acceleration input
\newcommand{\pRW}[1]{\prescript{W}{}{\pR{#1}}} 	% robot position wrt world frame (explicit)
\newcommand{\rotMatR}[1]{\rotMat_{R #1}}			% rotation matrix for robot orientation
\newcommand{\thrust}[1]{{f}_{R #1}}		% thrust
\newcommand{\inTorque}[1]{\vect{\tau}_{R #1}}		% thrust
\newcommand{\maxThrust}[1]{h_{#1}}				% maximum thrust
\newcommand{\thrustIJ}{\thrust{ij}}				% thrust robot ij
\newcommand{\maxThrustIJ}{\maxThrust{ij}}		% maximum thrust robot ij
\newcommand{\gravityIJ}{\vect{g}_{ij}}			% gravity vector robot ij
\newcommand{\massR}[1]{m_{R#1}}					% robot mass
\newcommand{\inertiaR}[1]{\vect{J}_{R#1}}		% robot inertia

% Admittance control
\newcommand{\dampingA}[1]{\matr{B}_{A#1}}		% damping
\newcommand{\springA}[1]{\matr{K}_{A#1}}		% spring
\newcommand{\inertiaA}[1]{\matr{M}_{A#1}}		% damping
\newcommand{\uA}[1]{\vect{u}_{A#1}}					% input
\newcommand{\paramA}[1]{\vect{\pi}_{A#1}}			% parameters

% System
\newcommand{\config}{\vect{q}}					% configuration system
\newcommand{\dconfig}{\vect{v}}			% derivative configuration system
\newcommand{\ddconfig}{\dot{\dconfig}}			% second derivative configuration system
\newcommand{\configR}{\config_R}					% configuration robots 
\newcommand{\dconfigR}{\dconfig_R}				% derivative configuration robots 
\newcommand{\ddconfigR}{\ddconfig_R}			% second derivative configuration robots
\newcommand{\configL}{\config_L}					% configuration load 
\newcommand{\dconfigL}{\dconfig_L}				% derivative configuration load 
\newcommand{\ddconfigL}{\ddconfig_L}			% second derivative configuration load
\newcommand{\state}{\vect{x}}						% state
\newcommand{\dynamicModelFun}{m}					% function that describes the dynamic model

% Equilibrium
\newcommand{\configEq}{\bar{\config}}			% configuration Load equilibrium
\newcommand{\configLEq}{\bar{\config}_L}		% configuration Load equilibrium
\newcommand{\configREq}{\bar{\config}_R}		% configuration Load equilibrium 
\newcommand{\paramAEq}[1]{\bar{\vect{\pi}}_{A#1}}			% parameters
\newcommand{\paramAEqInc}[1]{\hat{\bar{\vect{\pi}}}_{A#1}}
\newcommand{\paramAEqU}[1]{\hat{\bar{\vect{\pi}}}_{A#1}}% parameters
\newcommand{\pLEq}{\bar{\pos}_L}

\newcommand{\pLEquilib}{{\pos}^{eq}_L}% load position
\newcommand{\rotMatLEq}{\bar{\rotMat}_L}		% rotation matrix for load orientation
\newcommand{\paramASetEq}{{\Pi}_{A}(\configLEq)}				% parameters set for equilibrium
\newcommand{\paramASetEqPrime}{{\Pi}_{A}(\configLEq')}				% parameters set for equilibrium
\newcommand{\configRSetEq}{\mathcal{P}_{R}}			% robot pos set for equilibrium
\newcommand{\cableForceEq}[1]{\bar{\vect{f}}_{#1}}
\newcommand{\cableForceEqInc}[1]{\hat{\bar{\vect{f}}}_{#1}}				% cable force equilibrium
\newcommand{\cableForcesEq}{\bar{\vect{f}}}	
\newcommand{\cableForcesEqInc}{\hat{\bar{\vect{f}}}}					% cable forces equilibrium
\newcommand{\cableForcesSetEq}{\mathcal{F}_{L}}		% cable force set for equilibrium
\newcommand{\nullGrasp}{\vect{r}_L}				% orthogonal base of the null space of the grasp matrix
\newcommand{\internalTension}{t_L}				% internal tension
\newcommand{\internalForceDir}{\vect{n}_L}	% internal tension
\newcommand{\internalForceDirL}{\prescript{L}{}{\vect{n}}_L}	% internal tension in body frame
\newcommand{\pREq}[1]{\bar{\pos}_{R #1}}	
\newcommand{\pREqInc}[1]{\hat{\bar{\pos}}_{R #1}}		% robot position

\newcommand{\pREquilib}[1]{\pos^{eq}_{R #1}}
\newcommand{\configRparamASetEq}{\mathcal{S}(\configLEq)}
\newcommand{\configRparamAEq}{\bar{\vect{s}}}
\newcommand{\configRparamA}{{\vect{s}}}
\newcommand{\configSetEq}{\mathcal{Q}(\internalTension,\configLEq)}
\newcommand{\configSetEqZero}{\mathcal{Q}(0,\configLEq)}
\newcommand{\configSetEqZeroi}[1]{\mathcal{Q}_{#1}(0,\configLEq)}

\newcommand{\configSetEqPlus}{\mathcal{Q}^+(\internalTension,\configLEq)}
\newcommand{\configSetEqMinus}{\mathcal{Q}^-(\internalTension,\configLEq)}

\newcommand{\configRLSetEq}{\mathcal{R}(\internalTension,\configLEq)}
\newcommand{\configRLSetEqZero}{\mathcal{R}(0,\configLEq)}

\newcommand{\gr}[1]{\mathcal{G}}	

\newcommand{\error}[1]{\vect{e}_{#1}}	% position error with respect to the equilibrium robot position

% Static Analysis
\newcommand{\screwJacobian}{\matr{J}(\genCoord)}		% Screw Jacobian
\newcommand{\vectI}{\vect{v}_i}
\newcommand{\vectII}{{v}_{i,i}}
\newcommand{\minTensionResolving}{\underline{t}_i(\genCoord)}		% minimum tension on the i-th cable resolving W
\newcommand{\maxTensionResolving}{\overline{t}_i(\genCoord)}		% maximum tension on the i-th cable resolving W

% Stability
\newcommand{\stateEq}{\bar{\state}}
\newcommand{\invariantSet}{\Omega_{\alpha}}		% positively invariant set
\newcommand{\invariantSetZero}{\Omega_{0}}		% positively invariant set
\newcommand{\dVZeroSet}{\mathcal{E}}				% state subspace for which the dV is zero
\newcommand{\stateSetEq}{\mathcal{X}(\internalTension,\configLEq)}
\newcommand{\stateSetEqZero}{\mathcal{X}(0,\configLEq)}
\newcommand{\stateSetEqZeroi}[1]{\mathcal{X}_{#1}(0,\configLEq)}
\newcommand{\stateSetEqPlus}{\mathcal{X}^+(\internalTension,\configLEq)}
\newcommand{\stateSetEqMinus}{\mathcal{X}^-(\internalTension,\configLEq)}
\newcommand{\stateSetEqPlusPrime}{\mathcal{X}^+(\internalTension',\configLEq)}
\newcommand{\lyapunovFun}{V(\state)}				% lyapunov function

\newcommand{\Vadd}{V_R(\state)}				% lyapunov function
\newcommand{\dVadd}{\dot{V}_1(\state)}	

\newcommand{\dlyapunovFun}{\dot{V}(\state)}				% lyapunov function
\newcommand{\maxInvariantSet}{\mathcal{M}}

% Passivity
\newcommand{\inp}{\vect{u}}
\newcommand{\out}{\vect{y}}
\newcommand{\outputFunction}{\vect{\Phi}(\out)}
\newcommand{\viscous}[1]{c_{#1}}

\maketitle
\begin{abstract}
This work considers a large class of systems composed of multiple quadrotors manipulating deformable and extensible cables. The cable is described via a discretized representation, which decomposes it into linear springs interconnected through lumped-mass passive spherical joints. Sets of flat outputs are found for the systems. Numerical simulations support the findings by showing cable manipulation relying on flatness-based trajectories. Eventually, we present an experimental validation of the effectiveness of the proposed discretized cable model for a two-robot example. Moreover, a closed-loop controller based on the identified model and using cable-output feedback is experimentally tested.
\end{abstract}
\section{Introduction and Related Work}
 Deformable object manipulation is an important recent development in aerial robotics with potential applications ranging from fire fighting~\cite{kotaru2020multiple}, and in general the manipulation of fluid conduits~\cite{abhishek2021towards}, to waterway maintenance, e.g., floating litter collection \cite{gabellieri2023differential, gonzalezmorgado2025multirobotaerialsoft} or handling oil-spill events~\cite{kourani2021tethered}. Yet, for the challenges it involves~\cite{zhu2021challenges}, deformable object manipulation is still regarded as an open problem. An extensive survey on modeling, perception, and control techniques 
for deformable object manipulation is in~\cite{yin2021modeling}. This work especially focuses on the manipulation of cables through Uncrewed Aerial Vehicles (UAVs).

Aerial robotic manipulation has traditionally considered rigid cables or linear elastic cables, typically as tools to manipulate attached payloads. More recently, increasingly general models have been used to describe cable flexibility, too. 
In the following, we provide a quick overview of related works.
\subsubsection{Rigid cable models}
They neglect bending and length changes and typically involve manipulation of suspended payloads.
The differential flatness of a system composed of a single quadrotor and a slung load is shown in~\cite{sreenath2013trajectory}, while \cite{sreenath2013dynamics} extends the results to a point mass or rigid body manipulated by multiple UAVs.

\subsubsection{Inflexible extensible cable models}
 They neglect bendability but allow variations of the cable length.~\cite{kotaru2017dynamics} considers a payload suspended through a spring-damper cable below a single quadrotor, showing that the system is not differentially flat. Rigid bodies manipulated by multiple quadrotors are studied in~\cite{goodman2022geometric} and~\cite{goodman2021geometric}. Those works propose a controller based on a reduced model that neglects the cable elongations but shows good performance when applied to the elastic case. 
In~\cite{gabellieri2018study, gabellieri2020study}, elastic cables are considered in a cooperative manipulation scheme to regulate the pose of suspended rigid objects relying on individual UAV's force estimation. 
 
\subsubsection{Flexible inextensible cable models}\label{subs:f-i}
They allow cable bending but not length variations.
\cite{goodarzi2014geometric} proposes a controller to stabilize a quadrotor with a hanging bendable cable; similarly, an adaptive control to account for unknown mass is proposed in~\cite{dai2014adaptive}.
     \cite{goodarzi2016stabilization} considers cooperative aerial manipulation through flexible inextensible cables, and \cite{kotaru2018differential} shows the differential flatness of a point-mass or a rigid body suspended below multiple quadrotors through flexible inextensible cables.
     Moreover, the differential flatness of single- and multi-robot systems connected to these cables is shown in~\cite{kotaru2020multiple}.    A different modeling approach than is adopted in~\cite{d2021catenary,abhishek2021towards,gonzalezmorgado2025multirobotaerialsoft}, where two-quadrotors are connected by a cable modeled as a catenary curve or a parabola. Such models describe the kinematics of the cable through a small number of parameters but are mostly limited to quasi-static conditions.
     %%%%%%%%%%%
    \begin{figure*}[t]
    \centering
        \includegraphics[width=0.9\textwidth]{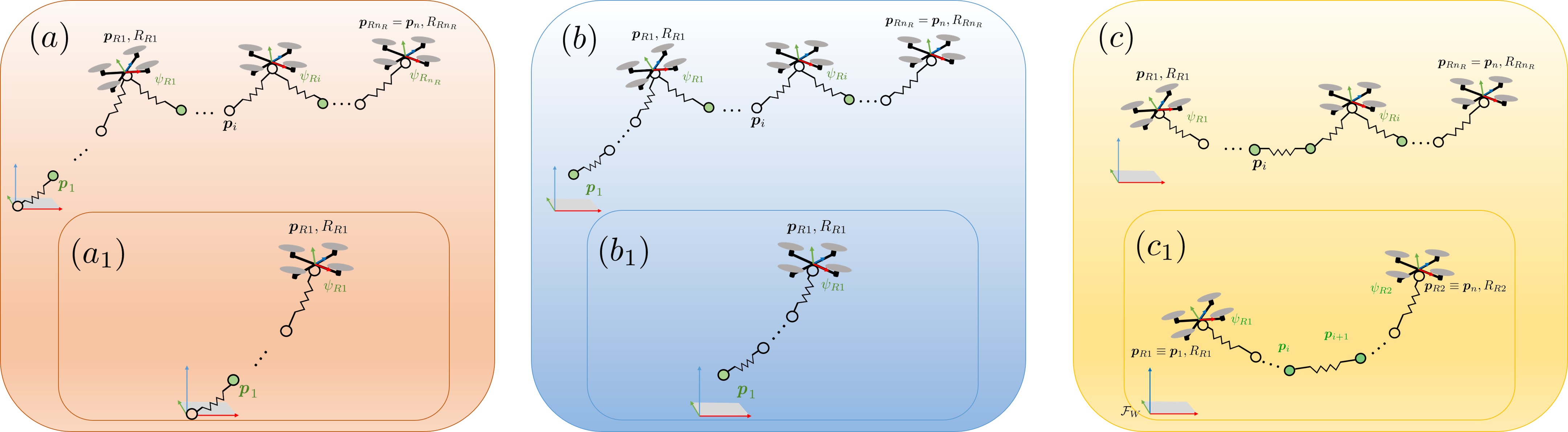}
    \caption{Schematics of the considered system composed of multiple quadrotors attached to an elasto-flexible cable.} \label{fig:all_sys}
\end{figure*}
%%%%%%%%%%%%%
\subsubsection{Flexible and extensible cable models}  They allow the cable to bend and change its length, thus enlarging the application range.  
 While the use of such models is found in manipulation with ground robots~\cite{pezzementi2008modeling, yin2021modeling, arriola2020modeling}, it is still mostly unexplored in the \textit{aerial} robotic manipulation domain. \cite{gabellieri2023differential} demonstrated the differential flatness of the discretized cable model composed of springs interconnected by lumped-mass passive spherical joints and attached between two quadrotors. The model incorporates external viscous forces, too, and it is the same used in this work. Other methods consider a distributed-parameter cable model, such as  \cite{liu2017modeling}, which applies back-stepping control to suppress the vibrations of a hose transported by a fixed-wing vehicle. To control the cable shape, \cite{shen2024aerial} applies proper orthogonal decomposition to the partial differential equations that describe the dynamics of the cable, combined with the ordinary differential equations that describe the dynamics of the attached quadrotor.   
 
 The main contributions of this work are as follows. \begin{itemize}
     \item It demonstrates the differential flatness of a large class of tethered multi-UAV systems that go beyond the 2-UAV case analyzed in \cite{gabellieri2023differential}. From a methodological point of view, compared to \cite{kotaru2020multiple}, the present work considers a more general model that includes the elasticity of the cable, which cannot be ignored in different real-world scenarios~\cite{1995-JosRah,goodman2022geometric}. The proposed discretized elastic model also allows finding meaningful flat outputs, namely, the positions of certain points on the cable. 
     \item Simulation results support the theoretical findings.
     \item Filling a gap left open in~\cite{kotaru2020multiple,gabellieri2023differential}, this work presents for the first time an experimental validation of the efficacy of the proposed discretized cable model in an aerial manipulation system. The parameters of the cable model are identified from real data. Hence, the identified model is tested in a series of new real experiments, characterized by diverse velocities and cable trajectories, performed using two UAVs. 
     \item A model-based manipulation strategy that closes the loop on the cable output is also tested in real experiments, showing the ability to reduce the output error in the presence of parameter uncertainties.
   \end{itemize}
\section{Model}\label{sec:modeling}
%Consider a system composed of a deformable and elastic cable manipulated by two quadrotor UAVs. 
Fig.~\ref{fig:all_sys} is a schematic representation of the different systems considered in this work.
Especially, \begin{itemize}
    \item systems $(a)$ are a class of systems in which one end of the cable is attached to the ground and the remaining part is suspended through a certain number of quadrotors attached at will along it; one quadrotor is attached to the other end of the cable. System $(a_1)$ is a subset of that class of systems, where there is only one robot.
    \item In class $(b)$, are free-flying systems where one end of the cable is hanging, while the other end is attached to the last quadrotor; systems $(b_1)$, composed of one quadrotor manipulating a hanging cable, are a subset of the $(b)$ class. 
    \item In class $(c)$ are systems in which each ending of the cable is attached to a quadrotor; $(c_1)$, where two quadrotors manipulate a cable from its extremities, is a widely studied subset of class $(c)$ \cite{gabellieri2023differential, d2021catenary}.\end{itemize}

We define an orthonormal inertial frame $\frame_W=\{ \origin_W , \vX_W, \vY_W , \vZ_W \}$, where $\origin_W$ is thew origin and $\vX_W, \vY_W , \vZ_W$ are three orthogonal unit vectors. We call $\vE{i}$ the $\ith{i}$ column of the 3-by-3 identity matrix, $\eye{3}$.

We consider deformable elastic cables and model them through a series of $n$ point masses interconnected by springs through passive spherical joints. Let us define $\mathcal{I}=\{1,\cdots,n\}$ the set of the indexes that we use to refer to the point masses. Hence, we indicate the position of mass $m_i$ expressed in $\frame_W$ as $\pos_i\in\nR{3}$, for  $i\in\mathcal{I}$. For class $(a)$, we also define as $\pos_0=\bm{0}$ the position of the end of the cable attached to the ground; without loss of generality, attached to $\origin_W$. As said, the masses are connected to elastic links modeled as linear springs; such a modeling approach for deformable objects has proven effective when dealing with small deformations~\cite{yin2021modeling}. For the spring that has one end attached to the point in position $\pos_j$ and the other to the one in position $\pos_{j+1}$ with $j\in\mathcal{I}\setminus \{n\}$ we define a stiffness coefficient $k_j\in\nR{+}$ and rest length $\length{j}\in\nR{+}$. The force exerted by the $\ith{j}$ elastic segment on the $\ith{j}$ point mass is indicated as $\cableForce{j}.$ Consequently,  $-\cableForce{j}$ is the force exerted by the same segment on the $\ith{(j+1)}$ mass. 
We assume that $\cableForce{j}\neq\vect{0}$. For systems in class $(a)$, we also define a spring with elastic coefficient $k_0\in\nR{+}$ and rest length $\length{0}\in\nR{+}$ that connects the first mass to the ground and exerts a force $\cableForce{0}$ on it.  Considering a simplified Hook's law to model the springs, we have that $\forall i\in\mathcal{I}\setminus \{n\}$,          
\begin{align}
    \cableForce{i}=-&\springCoeff{i}\left[\left(  \pos_{i} - \pos_{i+1}\right)-\length{i}\frac{\pos_{i} - \pos_{i+1}}{||\pos_{i} - \pos_{i+1}||}, \right]%=\nonumber\\
  % = & \springCoeff{i} \left[\left(  \pos_{i+1} - \pos_{i}\right)-\length{i}\frac{\cableForce{i}}{||\cableForce{i}||}\right]
  \label{eq:force}
\end{align}
where we indicated with $||*||$ the 2-norm.
Force $\cableForce{i}$ acts along the unit vector ${\frac{\pos_{i} - \pos_{i+1}}{||\pos_{i} - \pos_{i+1}||}= \frac{\cableForce{i}}{||\cableForce{i}||}.}$ Moreover,  for systems $(a)$ $\cableForce{0}=\springCoeff{0}\left[  \pos_{1}+\length{0}\frac{ \pos_{1}}{|| \pos_{1}||} \right]$, where $-\cableForce{0}$ acts on the mass $m_1$.

A set of $n_R$ quadrotor UAVs are attached through passive spherical joints to some of the masses along the cable. We indicate with $\mathcal{R}\subset \mathcal{I}$ the subset containing the indexes of the point masses directly attached to a robot, with $|\mathcal{R}=n_R$. We adopt the usual assumption \cite{tagliabue2017collaborative, goodman2022geometric} that the masses are attached to the UAVs' CoM so that the robot's attitude dynamics is decoupled from the rest of the system's dynamics. 
For $j\in\mathcal{R}$, we define the frame ${\frame_{Rj}=\{ \origin_{Rj},  \vX_{Rj}, \vY_{Rj}, \vZ_{Rj}\}}$ attached to the $\ith{j}$ cable point, coincident with the CoM of the corresponding UAV. %\textcolor{magenta}{AF: what about changing $i'$ in another index like $k, j, l, h, s, r$ or similar?}
The position of $\origin_{R_{j}}$ expressed in $\frame_W$ is indicated as $\pos_{Rj}$. The rotation matrix that expresses the attitude of $\frame_{Rj}$ w.r.t. $\frame_W$ is indicated as ${\rotMat_{Rj}\in{SO(3)}}$. For the quadrotor attached to the $\ith{j}$ point, %\textcolor{magenta}{AF: isn't $i'$ the index of the mass? so for example the 40-th quadrotor doesn't mean that there are 40+ quadrotors, right?}\textcolor{blue}{Exactly, the 40th quadrotor means it is attached to the discrete point number 40. The number of quadrotors is the cardinality of $\mathcal{R}$. if $\mathcal{R}=\{3,21,40\}$ there are 3 quads in positions 3, 21, 40. This is handy when writing the dynamics in compact form in (2a-2d). }\textcolor{magenta}{AF: ok bu then saying the `$\ith{i'}$ quadrotor' sounds strange to me because if for example $i'$ is 40 I interpret it as the 40-th quadrotor but actually there are no 40 quadrotors but only 3. Additionally I don't know how to define the next and previous quadrotor in the list if I need to. which one is the next to $i'$? it is not $i'+1$.}\textcolor{blue}{you're right! I changed the text saying the quadrotor attachded to the kth point instead} \textcolor{magenta}{To overcome these limitations you could   introduce a map $\rho$ from ${\cal I}_R$ to ${\cal R}=\{1,\ldots,n_R\}$, which assigns to each mass index which has an attached quadrotor an incremental  number, and then one can use the indexes in one or the other set depending on the needs.Feel free to ignore this suggestion of course.}
 we call $\yaw_{Rj}$ its yaw angle, ${m_{Rj}}$ its mass, and  ${\inertiaR{j}\in\nR{3\times 3}}$ its rotational inertia.  Its total thrust and torque, regarded as the control inputs, are indicated as ${\thrust{j}\in\nR{1}}$ and ${\inTorque{j}\in\nR{3},}$ respectively.

For $j\in\mathcal{R}$ and  $i\in\mathcal{I}\setminus\mathcal{R}$, and defining $\cableForce{0}=\bm{0}$ for classes $(b), (c)$, the dynamics equations of systems $(a)-(c)$ are as follows
%
% \begin{IEEEeqnarray}{CC}
% \IEEEyesnumber \label{eq:sys_dyn}\IEEEyessubnumber
% \bar{m}_{R1}\ddpos_{R1}=-\bar{m}_{R1}g\vE{3}+\cableForce{1}+\rotMat_{R1}\thrust{1}\vE{3}, \ \text{if}\  1\in\mathcal{R}  \IEEEyessubnumber \label{eq:dyn_1}\\
% \bar{m}_{R2}\ddpos_{R2}=-\bar{m}_{R2}g\vE{3}-\cableForce{n-1}+\rotMat_{R2}\thrust{2}\vE{3}\IEEEyessubnumber \label{eq:dyn_2} \\
% \inertiaR{j}\angAcc_{Rj}=-\angVel_{Rj}\times\inertiaR{j}\angVel_{Rj} +\inTorque{j}, \quad j=\{1,2\}\IEEEyessubnumber \label{eq:dyn_3}\\
% \dot{\rotMat}_{Rj}=\angVel_{Rj}\times\rotMat_{Rj}, \quad j=\{1,2\}\IEEEyessubnumber \label{eq:dyn_4}\\
% m_i\ddpos_{i}=-m_{i}g\vE{3}+\cableForce{i}-\cableForce{i-1}-\viscous{i}\dpos_i, \;\; 1 < i < n \IEEEyessubnumber \label{eq:dyn_5}
% \end{IEEEeqnarray}
\begin{IEEEeqnarray}{RL}
\IEEEyesnumber \label{eq:sys_dyn}\IEEEyessubnumber
\bar{m}_{Rj}\ddpos_{Rj}&=-\bar{m}_{Rj}g\vE{3}+\cableForce{j}-\cableForce{j-1}+\rotMat_{Rj}\thrust{1}\vE{3}  \IEEEyessubnumber \label{eq:dyn_posR}\\
\inertiaR{j}\angAcc_{Rj}&=-\angVel_{Rj}\times\inertiaR{j}\angVel_{Rj} +\inTorque{j} \IEEEyessubnumber \label{eq:dyn_rotR}\\
\dot{\rotMat}_{Rj}&=\angVel_{Rj}\times\rotMat_{Rj}\IEEEyessubnumber \label{eq:dyn_RR}\\
m_i\ddpos_{i}&=-m_{i}g\vE{3}+\cableForce{i}-\cableForce{i-1}-\viscous{i}\dpos_i, \IEEEyessubnumber \label{eq:dyn_mi}
\end{IEEEeqnarray}
where $\bar{m}_{Rj}=m_{Rj}+m_j$;  $\angVel_{Rj}, \angAcc_{Rj}\in\nR{3}$ are the angular velocity and angular acceleration of the quadrotor attached to the $\ith{j}$ point, respectively. The gravitational acceleration is indicated as $g$. $\viscous{i}\in\nR{+}$ is a viscous friction coefficient used for modeling a dissipative effect  (e.g., the dissipation caused by partial immersion of the cable in water).

Note that the considered system has a number of degrees of freedom equal to ${3n+3n_R,}$ three for the position of each of the $n$ masses three for the orientations of the $n_R$ quadrotors. Moreover, the number of inputs is $4n_R$, namely the total thrust intensity $\thrust{*}$ and a three-dimensional moment $\inTorque{*}$ for each of the $n_R$ quadrotors. We can therefore expect to find at most $4n_R$ flat outputs for the system. 

\section{Differential Flatness}\label{sec:flatness}
Differential flatness is a structural property of dynamical systems whose states and inputs are completely expressed as functions of the so-called \textit{flat outputs} of the system and their derivatives. Given ${\dot{\vect{x}}=\phi(\vect{x}, \vect{u})}$, with $\vect{x}$  the state vector and $\vect{u}\in\nR{m}$ the input vector, the system is differentially flat if there exists an $m-$dimensional output vector $\vect{y}=h(\vect{x}, \vect{u}, \dot{\vect{u}}, \ddot{\vect{u}}, \dots, \vect{u}\deriv{r})$ such that
$\vect{x}=\alpha(\vect{y}, \dot{\vect{y}}, \dots, \vect{y}\deriv{q})$ and $\vect{u}=\beta(\vect{y}, \dot{\vect{y}}, \dots, \vect{y}\deriv{q})$, with ${(\star)\deriv{i}}$ being the $\ith{i}$ time derivative and $ \alpha(), \beta()$ differentiable functions~\cite{rigatos2015differential}.  %and $\alpha(\star)$, $\beta(\star)$ smooth functions 

\begin{example}\label{thm_1}
For any system in class $(a)$ in Fig.~\ref{fig:all_sys}, $\cup_{j\in \mathcal{R}\setminus\{n\}}\{  \pos_{j+1}, \yaw_{j} \}\cup\{\pos_{1},\yaw_{n}\}$ % $\forall j\in\mathcal{R}\setminus\{n\}$, the set $\{\pos_{1}, \pos_{j+1}, \yaw_{j}, \yaw_{n}\}$ 
is a set of flat outputs. 
\end{example}

\begin{figure*}[t]
    \centering
    \subfloat[\label{fig:a1_sim}]{\includegraphics[trim=3.5cm 8cm 4cm 10cm, clip, width=0.25\linewidth]{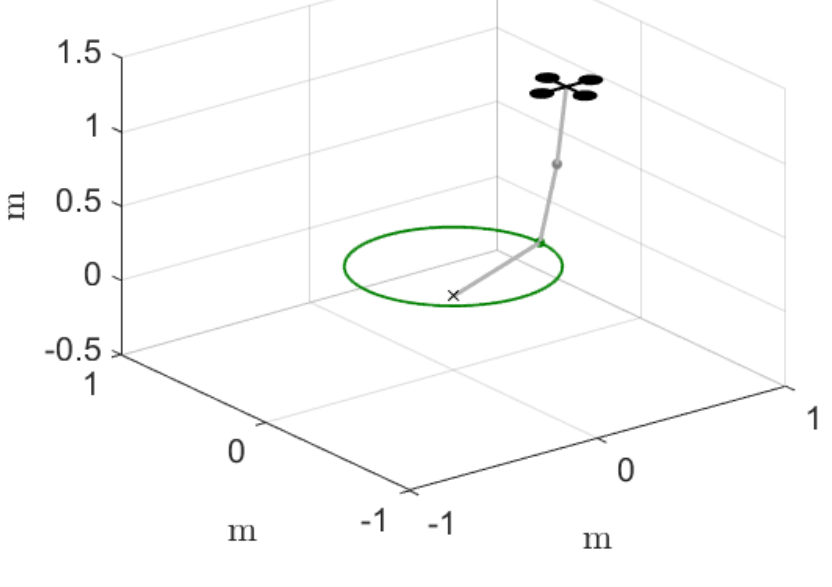}\includegraphics[width=0.25\linewidth, trim=3cm 8cm 3cm 8cm, clip]{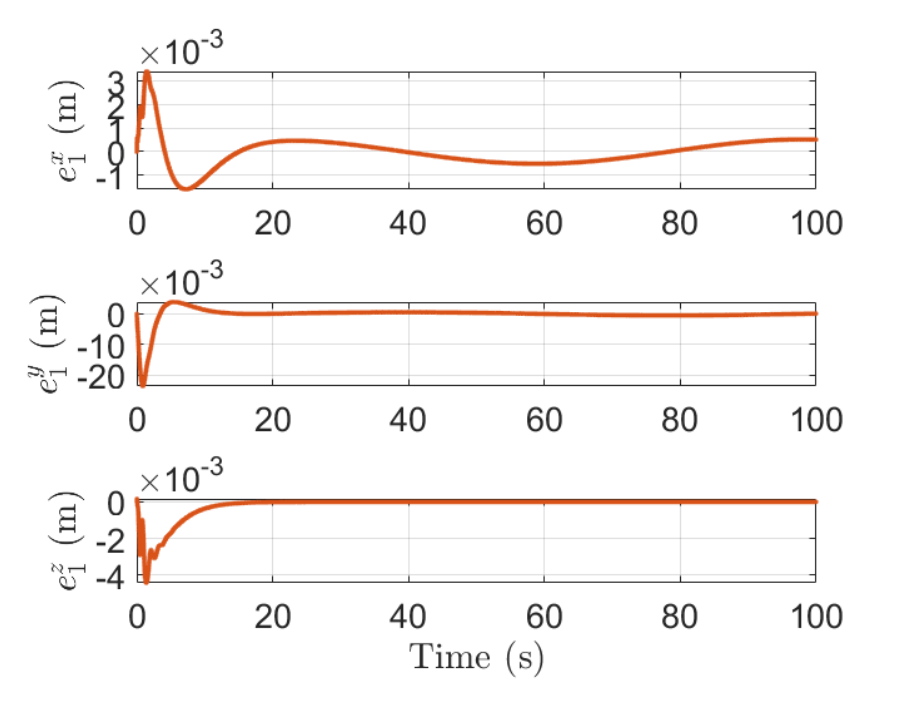}}
     \subfloat[\label{fig:b_sim}]{\includegraphics[ trim=3.5cm 8cm 4cm 10cm, clip,width=0.25\linewidth]{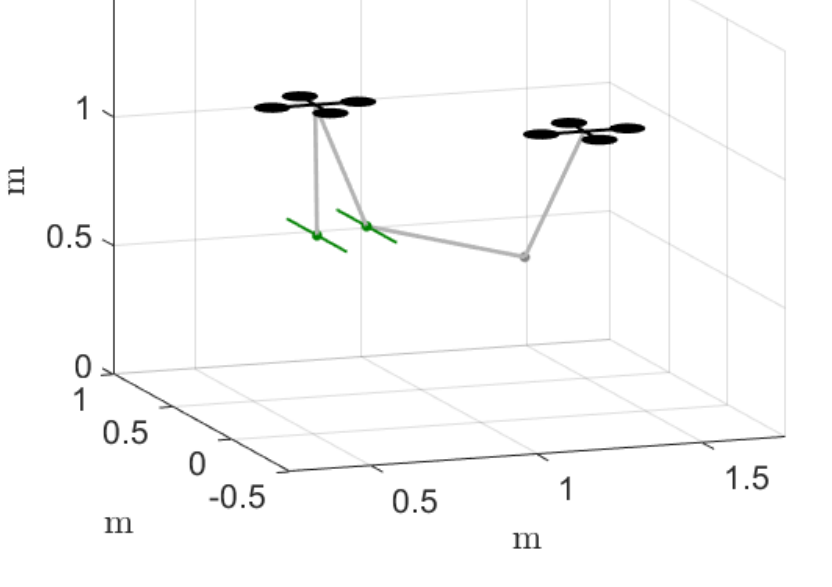}\includegraphics[ trim=3cm 8cm 3cm 8cm, clip, width=0.25\linewidth]{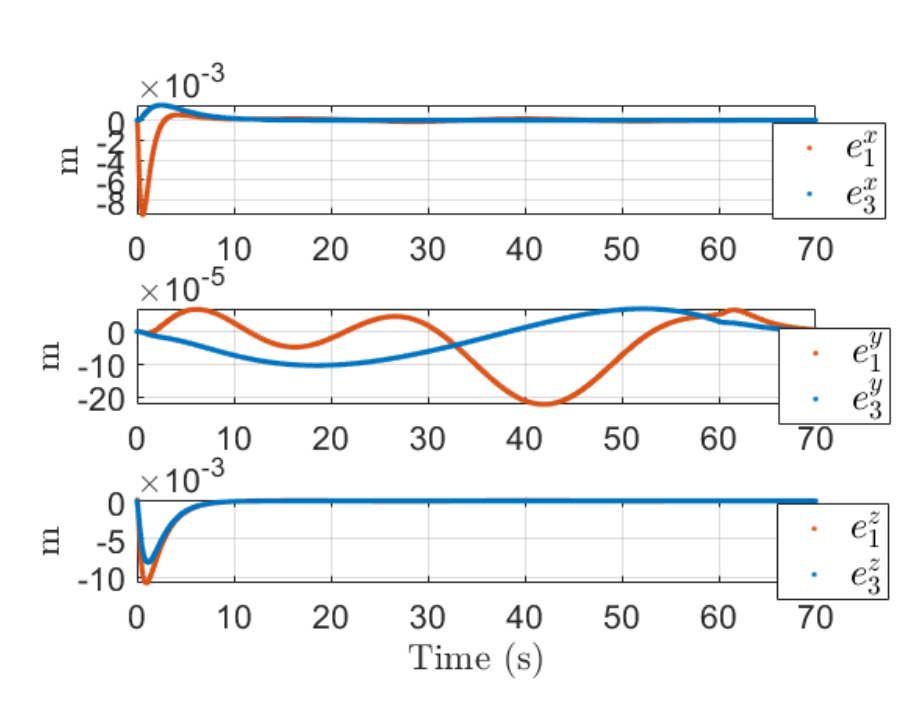}}
    \caption{Simulation results for a system of class $(a1)$ in (a) and class $(b)$ in (b). In the animation screenshot, the desired output trajectory and the output points are in green. In the plots, $x-, y-, z-$ coordinates of the output error.}
    \label{fig:sim_res}
\end{figure*}
%

%\textcolor{red}{AF: (The same comment applies to all the other results after.) expressed in this way it seems to me that the set of flat outputs is composed by two positions and two yaws, (8 outputs), i.e., that one can choose one of any quadrotors so that there are $n_R-1$ possible choices of flat outputs, but from the proof I understand that you need all the quadrotors positions (but the last one) and all the yaws. If you had defined the map above you could for example define the set of flat output as $\{\pos_{1},  \ldots,\pos_{\rho^{-1}(j)+1}, \yaw_{j}, \ldots ,   \yaw_{n}\}$ with $j=1,\ldots, n_R-1$, which to me immediately looks like a unique set with indeed $4n_R$ outputs, including all the quadrotors. An alternative is $\cup_{j=1}^{n_R-1}\{  \pos_{\rho^{-1}(j)+1}, \yaw_{j} \}\cup\{\pos_{1},\yaw_{n}\}$, or if you want to stick to your notation $\cup_{j\in \mathcal{R}\setminus\{n\}}\{  \pos_{j+1}, \yaw_{j} \}\cup\{\pos_{1},\yaw_{n}\}$. }\textcolor{blue}{right! thank you for this comment, I changed the result using the last notation you suggested.}

\textit{Remark} Result \ref{thm_1} tells us that the position of the hanging ending of the cable and of the points in the cable corresponding to the first masses after each quadrotor, together with the yaw of the robots, is a set of $3n_R+n_R$ flat outputs. In the work presented in~\cite{kotaru2017dynamics}, it is rightly noted how a spring-damper cable is not a flat system. Instead, we show how external viscous damping lets the system be flat; we experimentally support the proposed model choice in Section \ref{sec:exp}.

%\textcolor{red}{Maybe the theory holds only when considering stretched cables, while when the cable become being compressed, i.e., it is slack, not anymore: The functions of the output must be smooth which is the case only in that working region (however,~\cite{kotaru2020multiple} uses same functions to show flatness without discussing this, so might be I am wrong and there's no need). I could add a sentence to see that we are not considering slack cable and that is left as future work. }

\begin{proof}
\begin{enumerate}
\item \label{st:1} First, the force in the cable segment attached to the ground is a function of the flat outputs alone: \begin{equation}\label{eq:f0}
    \cableForce{0} = k_0\left(\pos_1-\length{0}\frac{\pos_{1}}{|| \pos_{1}||}\right).\end{equation}
    \item \label{st:2}Substituting \eqref{eq:f0} into \eqref{eq:dyn_mi}, one obtains
    \begin{align}\label{eq:f1}
& \cableForce{1}=m_i\ddpos_{1}+m_{1}g\vE{3}+k_0\left(\pos_1-\length{0}\frac{\pos_{1}}{|| \pos_{1}||}\right) +\viscous{1}\dpos_1.
    \end{align} Let us refer to the expression on the right-hand side of \eqref{eq:f1} as $\cableForceEq{1}$. That is a function of the flat outputs, and its derivatives are a function of the flat output derivatives.
    \item \label{st:3} Applying \eqref{eq:force} for $i=1$, one finds another expression of $\cableForce{1}$ depending on the known quantity $\pos_{1}$ and the unknown $\pos_{2}$. Combining that and \eqref{eq:f1}, one has
    \begin{equation}\label{eq:f1_2}
         \cableForceEq{1}=-\springCoeff{1}\left[\left(  \pos_{1} - \pos_{2}\right)-\length{1}\frac{\pos_{1} - \pos_{2}}{||\pos_{1} - \pos_{2}||}. \right]
    \end{equation}
    We note that the unit vector expressing the direction of the force is $\frac{\pos_{1} - \pos_{2}}{||\pos_{1} - \pos_{2}||}=\frac{\cableForceEq{1}}{||\cableForceEq{1}||}$. Then, we rewrite \eqref{eq:f1_2} as
\begin{equation}\label{eq:f1_3}   \pos_{2}=\pos_{1}+\frac{\cableForceEq{1}}{\springCoeff{1}}-\length{1}\frac{\cableForceEq{1}}{||\cableForceEq{1}||} ,
    \end{equation}
    which is now an equation in the only unknown $\pos_{2}$. We retrieve $\pos_{2}$ as a function of the flat outputs from \eqref{eq:f1_3}, and its derivatives are a function of the flat output derivatives by differentiating $\eqref{eq:f1_3}$.
\item \label{st:4} Repeat the previous steps \ref{st:2}-\ref{st:3} for increasing indexes, iteratively computing the expressions of the cable forces and the mass positions as a function of the flat output $\pos_{1}$ and its derivatives, until you reach an index $j\in\mathcal{R}$.
\item \label{st:5} When you arrive at $j\in\mathcal{R}$, you have $\pos_{j}=\pos_{Rj}$ and
\begin{align}\label{eq:mR_st5}
\bar{m}_{Rj}\ddpos_{Rj}=-\bar{m}_{Rj}g\vE{3}+\cableForce{j}-\cableForceEq{j-1}+\rotMat_{Rj}\thrust{j}\vE{3}, 
    \end{align}
    where $\pos_{Rj}$ and $\cableForceEq{j-1}$ are known as functions of the flat outputs and their derivatives from the application of previous steps \ref{st:3} and \ref{st:2}, respectively. Using the fact that $\pos_{j+1}$ is  a flat output by hypothesis, one finds $\cableForce{j}$ as a function of the flat outputs and derivatives as follows
    \begin{equation}\label{eq:fj_st5}
        \cableForce{j}=-\springCoeff{j}\left[\left(  \pos_{j} - \pos_{j+1}\right)-\length{j}\frac{\pos_{j}-\pos_{j+1}}{||\pos_{j}-\pos_{j+1}||}, \right]:=\cableForceEq{j}
    \end{equation}
    Substituting \eqref{eq:fj_st5} into  \eqref{eq:mR_st5}, one finds the only unknown, namely the thrust of the quadrotor $\rotMat_{Rj}\thrust{j}\vE{3}.$ From that and the flat outputs, which inlcude the yaw $\yaw_{Rj},$ and keeping in mind that  $\pos_{Rj}$ has been previously retrieved, the rotational matrix $\rotMat_{Rj},$ the moments $\tau_{Rj},$ and the angular velocity $\omega_{Rj}$ are computed as in \cite{mellinger2011minimum}.
\item \label{st:6} The previous steps  are repeated until index $n$ is reached.
\end{enumerate}
\end{proof}
\begin{example}\label{thm_2}
For any system in class $(b)$ in Fig.~\ref{fig:all_sys}, $\cup_{j\in \mathcal{R}\setminus\{n\}}\{  \pos_{j+1}, \yaw_{j} \}\cup\{\pos_{1},\yaw_{n}\}$ is a set of flat outputs. \end{example}

\begin{proof}
    The proof follows the same steps as in the one of Result \ref{thm_1}, this time with  $\cableForce{0}=\bm{0}$.
\end{proof}
\begin{example}\label{thm_3}
For any system in class $(c)$ in Fig.~\ref{fig:all_sys}, $\forall{i\in\mathcal{I}\setminus\{n\}}$, $i, i+1\notin \mathcal{R},$  $\cup_{j\in \mathcal{R}\setminus\{n\}, j>i}\{  \pos_{j+1}, \yaw_{j} \}\cup_{k\in \mathcal{R}\setminus\{1\}, k<i}\{  \pos_{k-1}, \yaw_{k} \}\cup\{\pos_{i},\pos_{i+1}, \yaw_n\, \yaw_{1}\} $
    %\item $\forall{i\in\mathcal{I}\setminus\{n\}}$, $i\in\mathcal{R}, i+1\notin \mathcal{R},$  $\cup_{j\in \mathcal{R}\setminus\{n\}, j>i}\{  \pos_{j+1}, \yaw_{j} \}\cup\{\pos_{i},\pos_{i+1}, \yaw_n\, \yaw_{1}\}\cup_{k\in \mathcal{R}\setminus\{1\}, k\leq i}\{  \pos_{k-1}, \yaw_{k} \} $
    %\item $\forall{i\in\mathcal{I}\setminus\{n\}}$, $i\notin\mathcal{R}, i+1\in \mathcal{R},$  $\cup_{j\in \mathcal{R}\setminus\{n\}, j\geq i+1}\{  \pos_{j+1}, \yaw_{j} \}\cup\{\pos_{i},\pos_{i+1}, \yaw_n\, \yaw_{1}\}\cup_{k\in \mathcal{R}\setminus\{1\}, k< i+1}\{  \pos_{k-1}, \yaw_{k} \} $

 is a set of flat outputs. \end{example}
%\textcolor{red}{AF: in this result I dont understand what is $i$, perhaps is $i\in\cal I$ and not $j$? but I am not sure, because also of the reason in the other comment before I am confused.}\textcolor{blue}{changed. I also changed the following Remark.}

\begin{proof}
    The proof that two consecutive points ($i$ and $i+1$) along subsystem $(c_1)$ are flat outputs is found in \cite{gabellieri2023differential}. Following that, the positions of the two quadrotors at the extremities of a cable portion are both found as functions of the flat outputs or their derivatives, so one can apply step \ref{st:5} in the proof of Result \ref{thm_1} in the direction of increasing and decreasing indexes and then apply steps \ref{st:2}-\ref{st:5} of the same proof recursively until the extreme quadrotors are reached.
\end{proof}
\textit{Remark} Result \ref{thm_3} tells that any two consecutive points along the cable and the position of the point mass after each quadrotor along the direction of the increasing indexes and any point mass position before each quadrotor along the direction of decreasing indexes, together with the robots' yaw angles are flat outputs. An example of flat outputs for the different systems is highlighted in green in Fig. \ref{fig:all_sys}. Result \ref{thm_3} considers the case where $i, i+1\notin\mathcal{R}$. This hypothesis can be easily relaxed considering either $i$, $i+1$, or both (such a case may be of poor practical interest) in $\mathcal{R}$. That has been avoided here for the sake of readability.
\subsection{Trajectory Generation}\label{sec:ctrl}
Based on the differential flatness, we can compute dynamically feasible trajectories for the system. We consider now the problem of letting the output  follow an assigned trajectory, indicated with the superscript $()^d$. For instance, for class (a) systems, we will have, $\forall j\in\mathcal{R}\setminus\{n\}$, $\pos_{j+1}^d$, $\pos_{1}^d$, $\psi_j^d,$ and $\psi_n^d$.
By applying the same steps showcased in the previous part of this section, one retrieves the whole system's desired state.
In practice, it is sufficient to compute all the forces in the cable and the positions of all the point masses over time, including the quadrotors'. Then, the control of the quadrotors along the desired smooth trajectories is done using state-of-the-art position controller~\cite{lee2010geometric}.

\section{Simulation Results}\label{sec:sim}
In this section, simulation results are presented to show the flatness-based trajectory generation.
The systems has been simulated using Matlab-Simulink, including the quadrotors with their under-actuated dynamics controlled via standard geometric
trajectory controller~\cite{lee2010geometric}. We present here simulation results for examples of systems in categories $(a)$ and $(b)$, while an instance belonging to class $(c)$ is extensively treated in the experimental section Section~\ref{sec:exp}.

In Fig.~\ref{fig:a1_sim}, are the results from a system in category $(a1)$ with $n=3$, where the quadrotor trajectory is generated for the output point to follow a circular trajectory of radius ${0.460\rm{m}}$ at a frequency of 0.08 Hz. The system is initialized such that the output is on the circle, but with an initial zero speed, hence, an initial velocity error.

Fig. \ref{fig:b_sim} contains simulation results from a system in category $(b)$ where $n=5.$ The system is initialized in $\pos_{1}=[0.6;0;0.697]\rm{m}$, $\pos_{R2}=[0.6;0;1.2]\rm{m}$, $\pos_{3}=[0.75;0;0.7193]\rm{m}$, $\pos_{4}=[1.251;0;0.7193]\rm{m}$, and $\pos_{R5}=[0.6+0.8;0;1.2]\rm{m}.$ The desired trajectory is a $\ith{5}$-order polynomial trajectory for the two outputs to move by ${0.5\rm{m}}$ along $\vY_W$ in 60 seconds. An initial error affects $\pos_{R5}$.

 Small errors are visible in the output tracking, due to UAV tracking errors, initial condition errors, and numerical errors: note that the proposed trajectory-generation method requires subsequent differentiations of non-linear functions, the number of which increases with the discretization of the cable; low-pass filtered numerical differentiation in Matlab-Simulink was used to obtain part of the needed quantities. The tracking error was observed to be influenced by the parameters of the differentiation block.
 \begin{figure}[t]
    \centering
   \subfloat[In black, 3D reconstruction of the real data ($\star^{meas}$) used for identification; in blue ($\star^{sim}$), identified model when subject to the same inputs, i.e., the positions of the extremities of the cable (red).     \label{fig:est_plots}]{\includegraphics[width=0.9\linewidth, trim=0cm 0cm 0cm 0cm, clip]{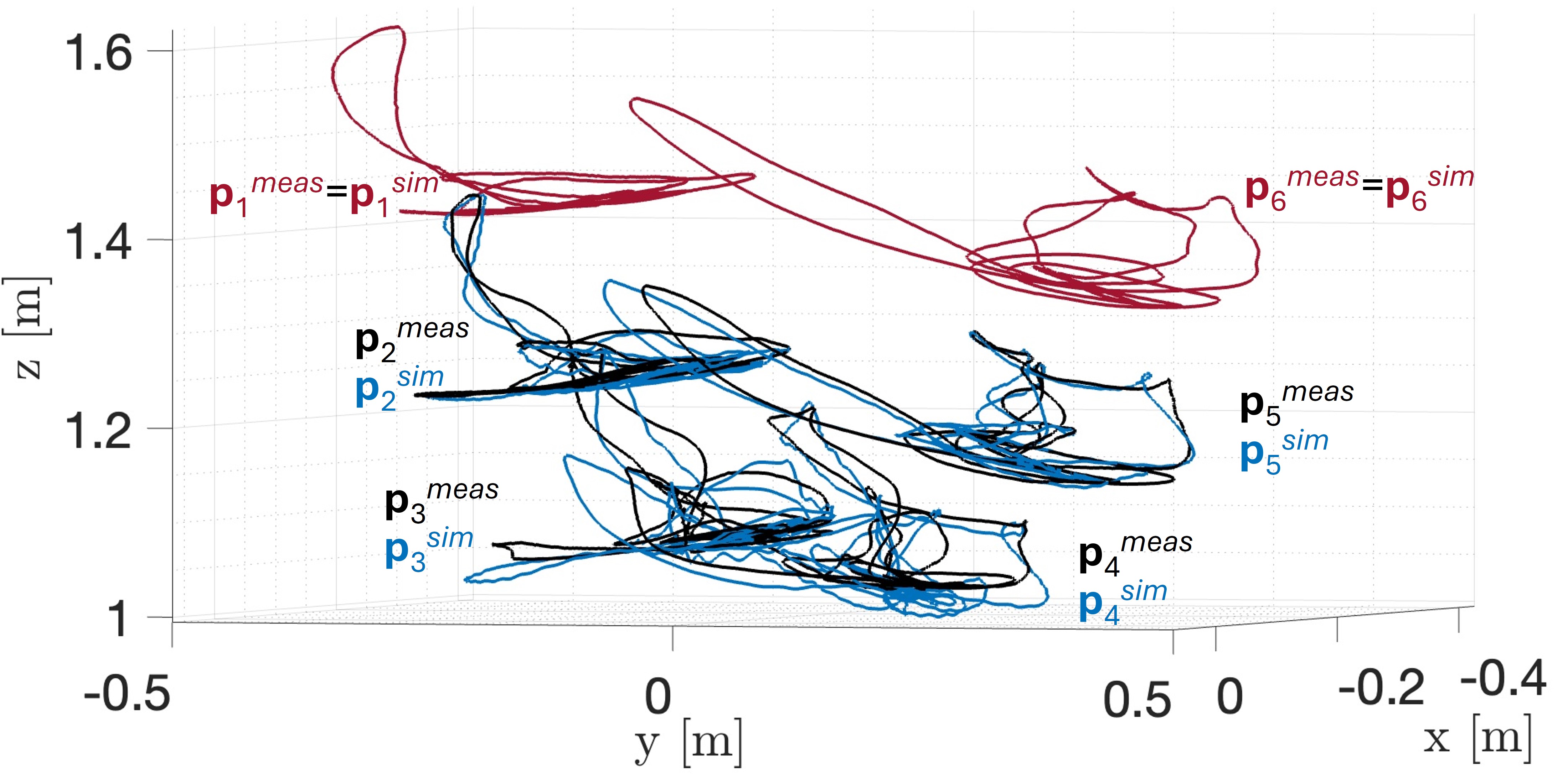}} \\
   \smallskip
   \subfloat[Error between the estimated and the measured positions of the four cable masses on the parameter identification dataset.
    \label{fig:est_err}]{\includegraphics[width=0.9\linewidth, trim=0cm 0cm 0cm 0cm, clip]{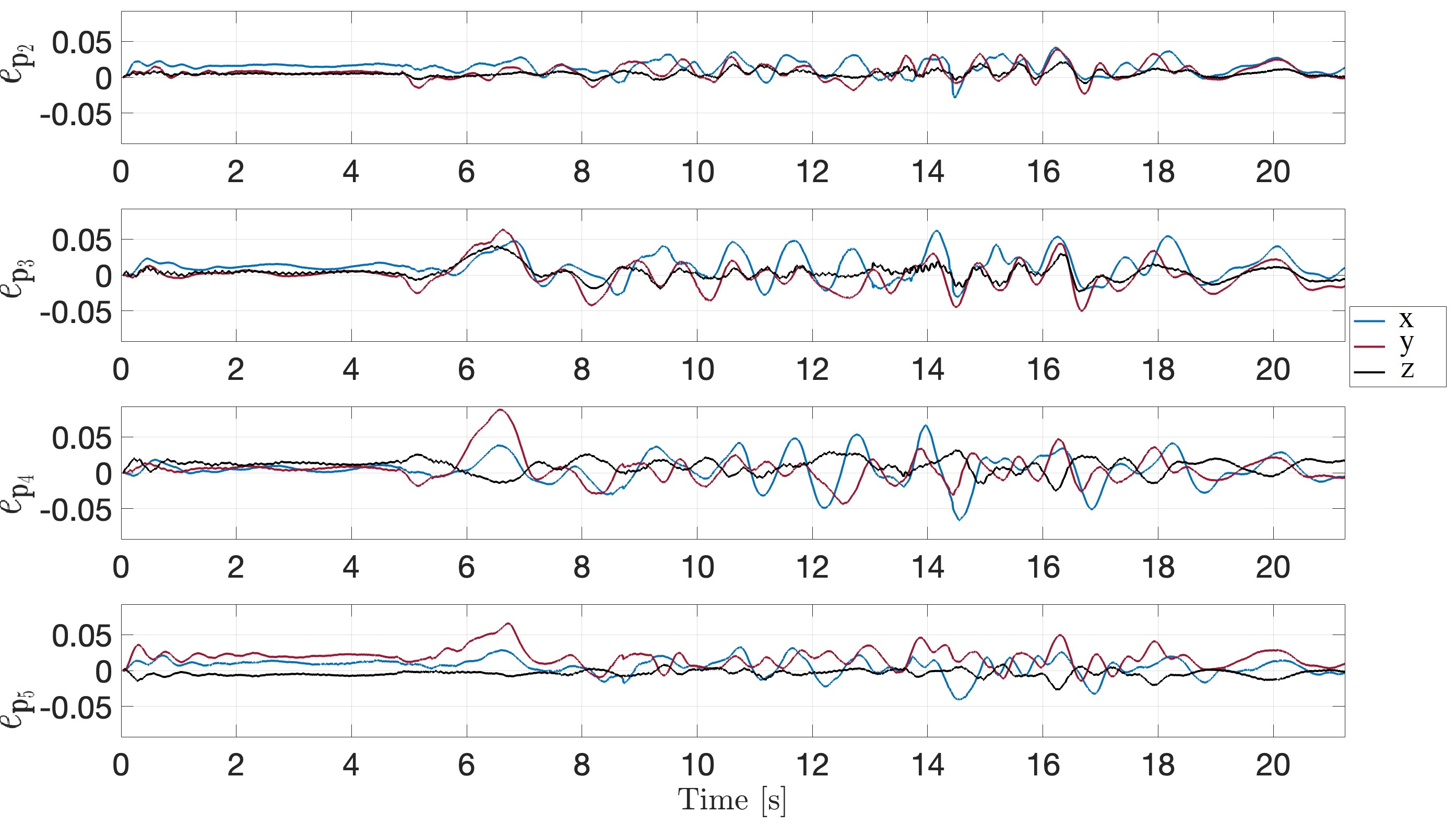}}\caption{Parameter identification results.}
\end{figure}
\begin{figure*}[!h]
    \centering
\includegraphics[width=\linewidth]{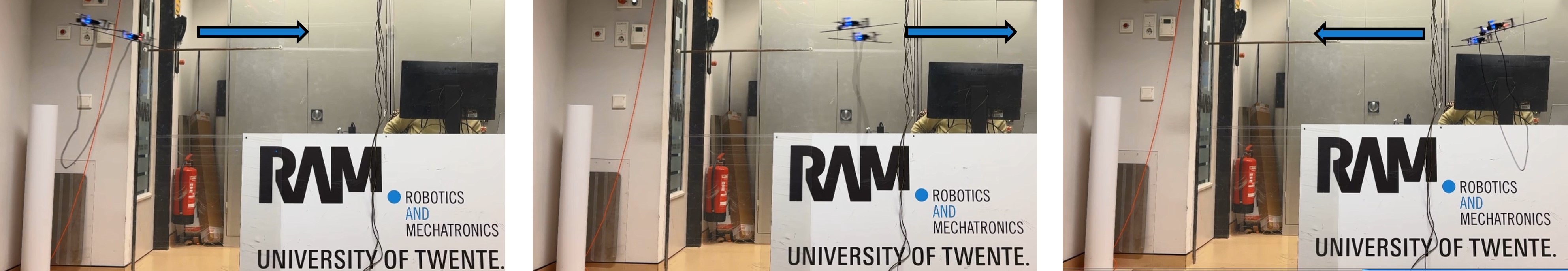}
    \caption{Three consecutive instants of a Test A experiment.}
    \label{fig:exp_1}
\end{figure*}
\begin{figure*}[!h]
\centering
\includegraphics[width=0.4\textwidth,height=3cm]{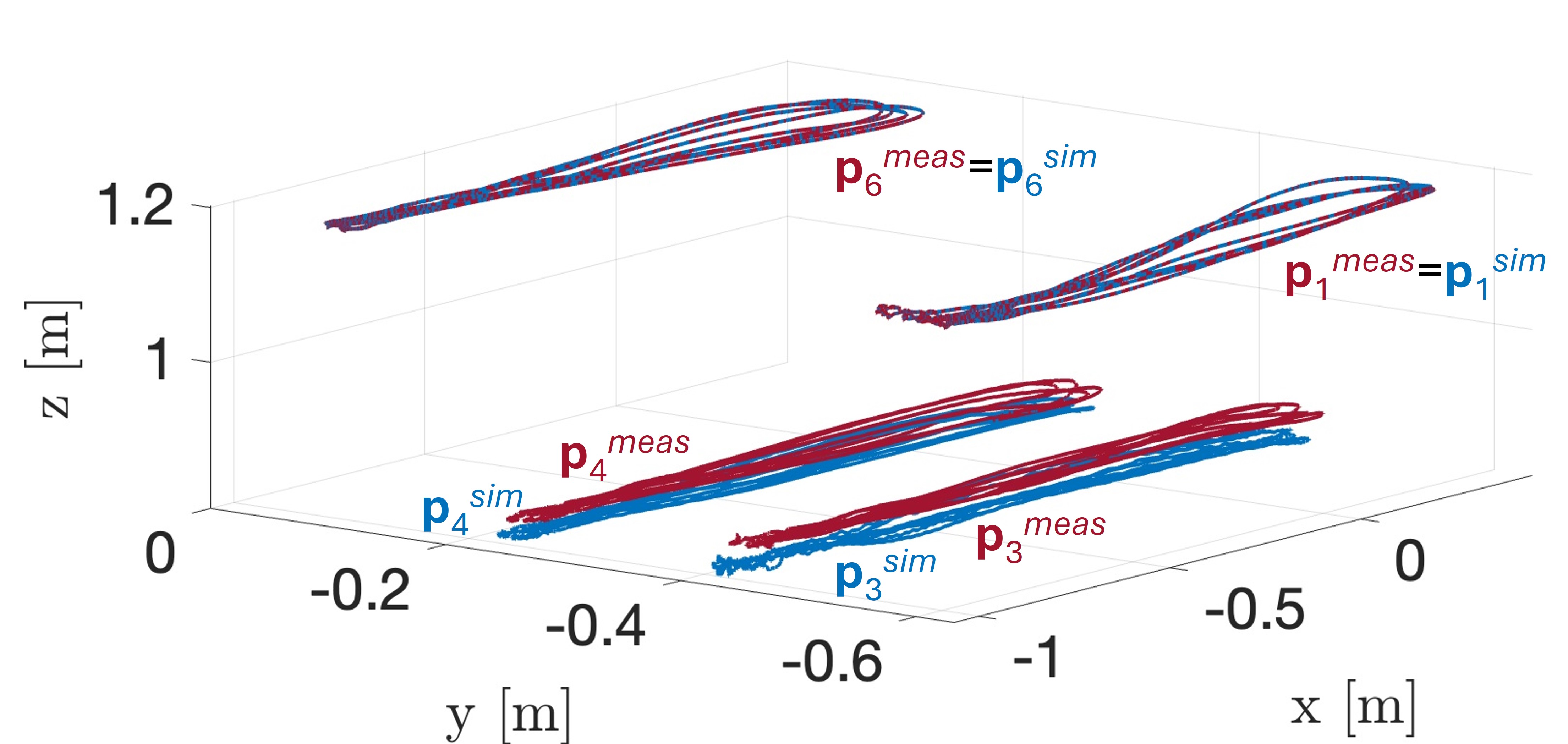} \qquad \qquad \includegraphics[width=0.4\textwidth,height=3cm]{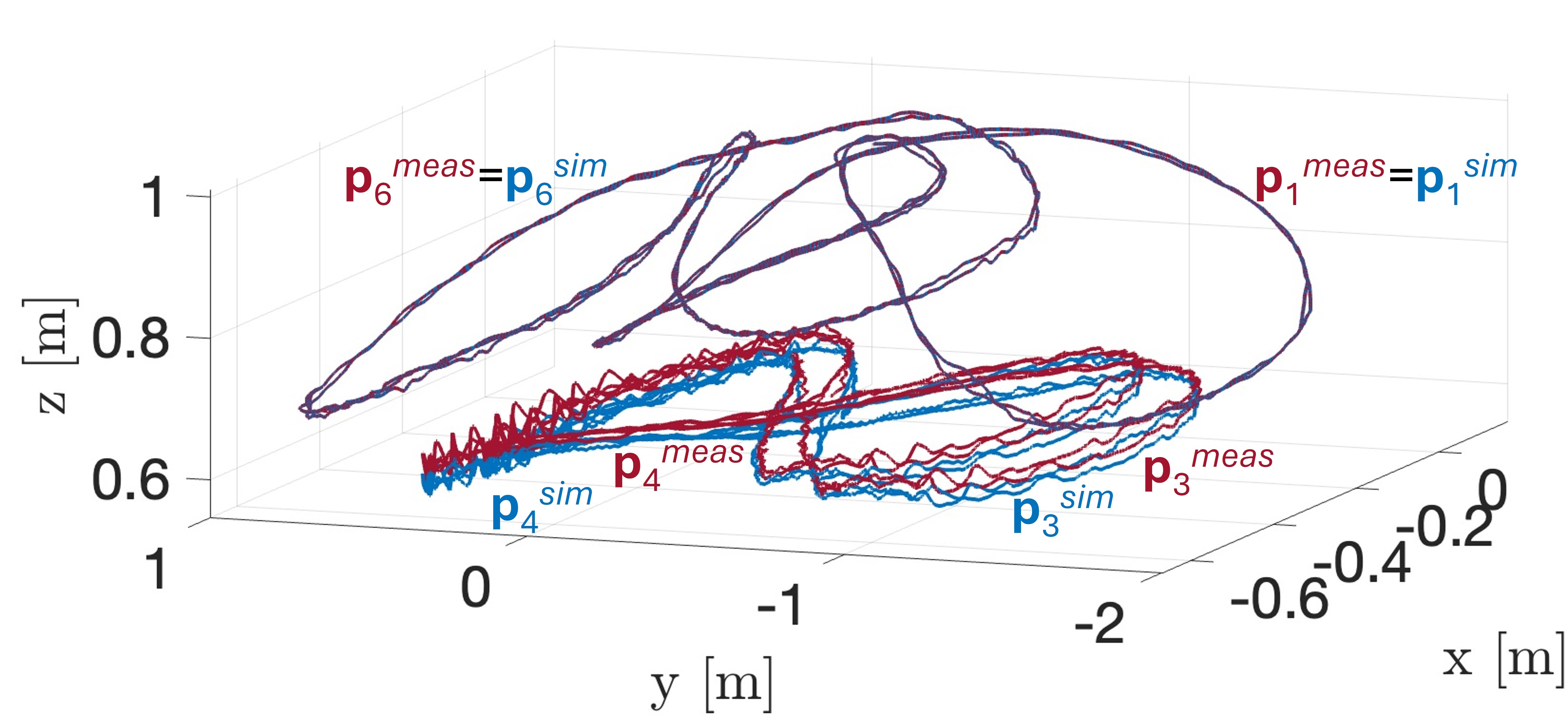}\\ \quad
\includegraphics[width=0.40\textwidth, height=2.5cm]{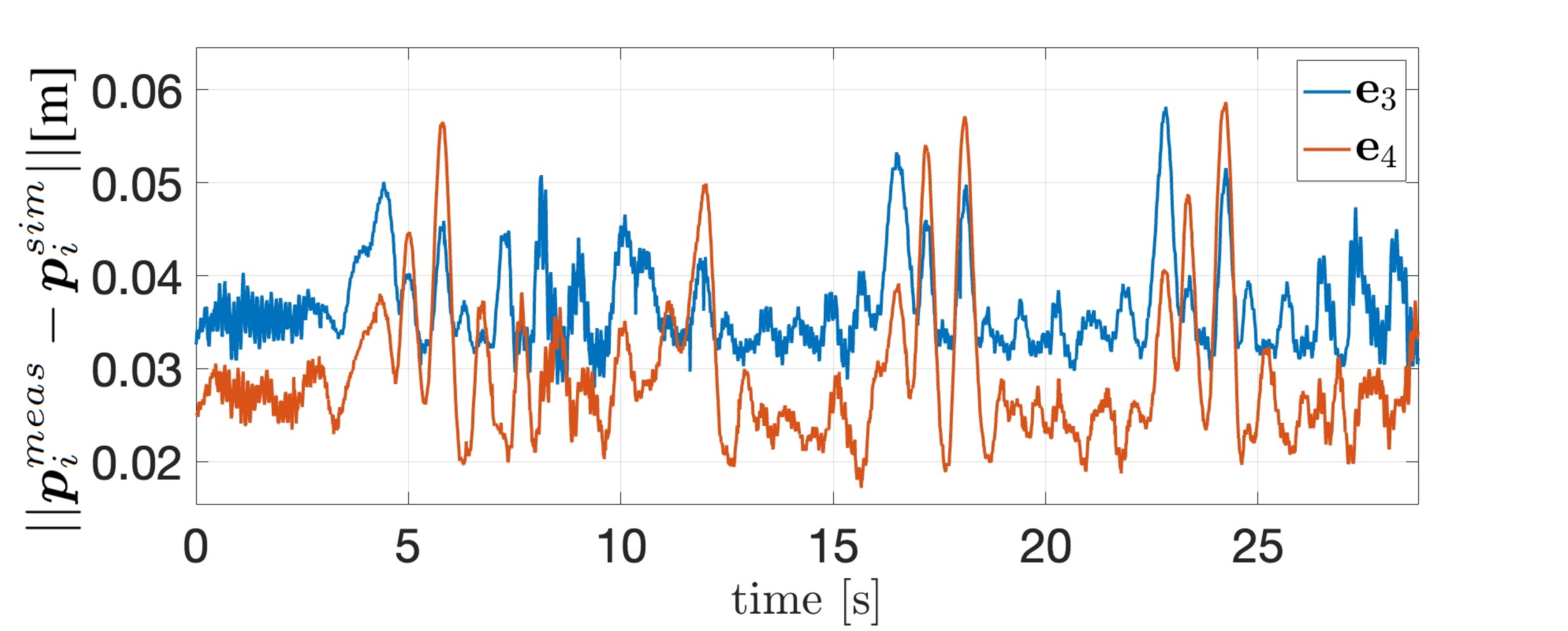} \qquad\qquad \includegraphics[width=0.40\textwidth,height=2.5cm]{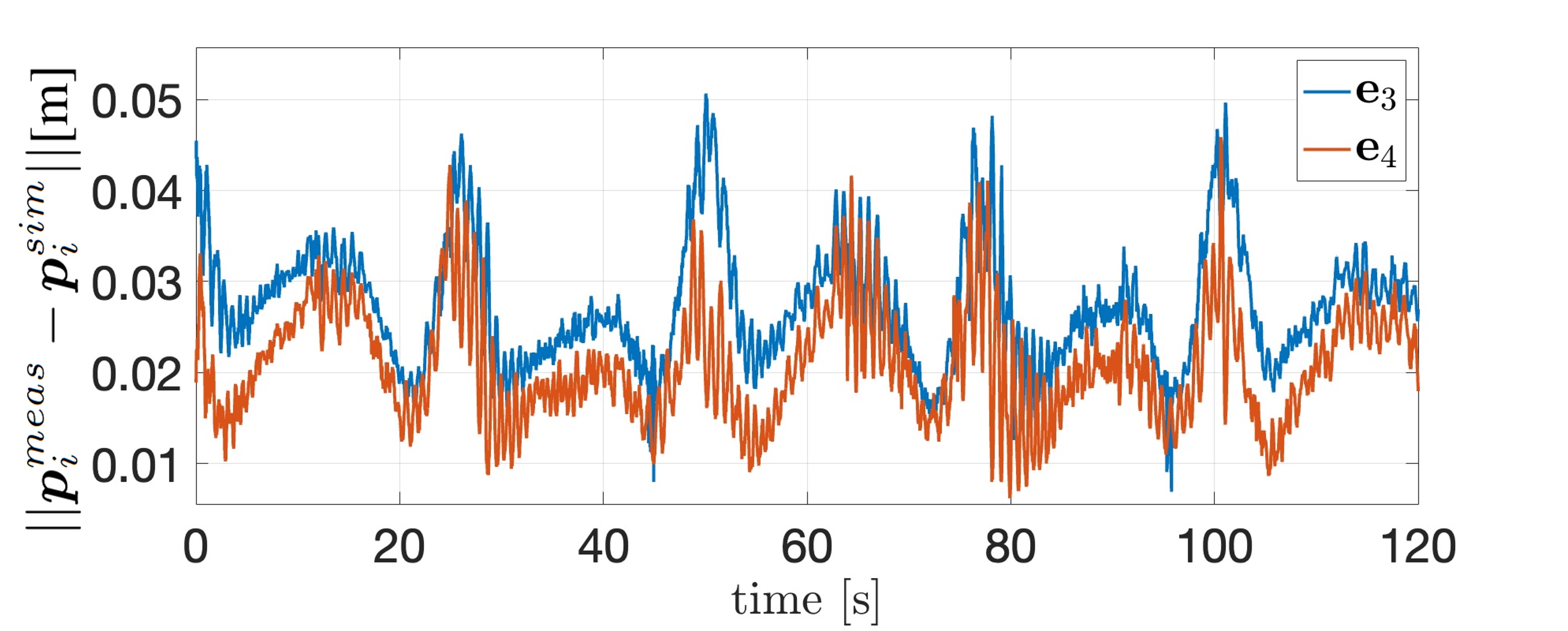}
    \caption{$||\bm{p}_i^{meas}-\bm{p}_i^{sim}|| $[m] Top: measured (red) and estimated (blue) cable trajectories subject to the same quadrotor trajectories for the narrow fast Test A (left) and the fast Test B (right). Below are the corresponding output error plots, where $\bm{p}_i^{sim}$ is the estimated value of the output $\bm{p}_i$ and $\bm{p}_i^{meas}$ the measured one. $||.||$ is the 2-norm  }\label{fig:open_loop_tests}
\end{figure*}
%%%%%%%%%%%%%%
\section{Experimental Results}\label{sec:exp}
\subsection{Cable Parameter Identification}
The proposed elastic and flexible cable model allows the convenient identification of controllable outputs and the generation of dynamically feasible trajectories. However, the question remains whether such a model describes the behavior of a real cable accurately enough.  To answer this question, this section describes the cable parameter identification and shows results comparing the behavior of the ideal model's and the actual cable's outputs for a class $(c_1)$ system. 

In the experiment, a 1-meter cable with mass $7\cdot 10^{-3}$ kg has been uniformly discretized into 6 points, including the two extremes where the robots would be attached. Hence, $n=6$ and $n_R=2$. Reflective tape has been attached to each point mass to record their position using a motion capture system; the data was recorded at 100 Hz. No quadrotors were used for the collection of the parameter identification dataset. While that may result in some unmodeled components such as aerodynamic interference of the propellers on the cable, it allows for a wider range of motion of the cables without worrying about instability issues and avoids the need for planning complex robot trajectories, which may result in being unfeasible. Instead, the cable was manually perturbed via the extremities. As the variability of the dataset affects the parameter identification, a wide range of configurations and velocities have been generated. Validation with the robots at the extremities of the cable is shown in the last part of this section.  

 Under the assumption of a uniform mass distribution along the cable, the mass of each point 
 was considered known and equal to the total mass divided by the number of discrete points. To reduce the dimension of the parameter space, the rest length of each spring has been considered known and equal to the average value of the distance between the two spring extremities over the whole dataset (see Table \ref{table:param}). Moreover, the viscous friction coefficient has been assumed the same for all points, $c_i=c$.
The vector of unknown parameters is $\bm{\theta}\in\nR{6}$, containing the spring and viscous coefficients. Note that even for not particularly elongating cables, the elasticity in our discretized model allows capturing the variable distance between consecutive points during the task execution; that is especially present when choosing coarse discretizations, which, on the other hand, are preferable in terms of the computational cost of the trajectory generation. 

Define the state vector $\bm{x}\in \nR{6(n-2)}$ containing the positions and velocities $\pos_{i}, \dpos_{i}$ for $i\in\mathcal{I}\setminus\mathcal{R}.$ Its estimate is defined as $\hat{\bm{x}}$. The parameter identification method is based on the minimization of the error between the measured state over time and the estimated state; the latter is obtained by integrating the cable dynamics subject to the initial conditions and inputs of the real system. $\ith{4}$-order Runge-Kutta integration method has been used to solve the dynamics. The measured positions of the two extreme points have been considered as the inputs at this stage. 
To improve the quality of the estimation result,
homotopic-based convexity has been used as introduced in \cite{bonilla2010convexity}. That method introduces a homotopy parameter $\lambda\in(0,1)$, used to move across the solution space, and an auxiliary state $\bm{x}_c$. $\lambda$ is decreased during the optimization execution to find a better solution to the non-convex optimization problem (closer to the global optimum) under the assumption that the initial cost function is affine in the parameters (as in our case with our unknown parameter selection). The auxiliary state is obtained using the measured state at the \textit{previous} time step as the initial condition. Conversely, the state estimate $\hat{\bm{x}}$ uses the measured initial condition only at the initial time step.
The method results in the following cost function minimization
\begin{align} \label{eq_mbpi:obj}
\begin{split}
    &\underset{\boldsymbol{\theta},\hat{\mathbf{x}}_i\hat{\mathbf{x}}_{i_c}}{\text{min}} \sum_{i=2}^{n-1} \left( \sum_{t=0}^{t_{\text{end}}} 
    \frac{1}{\lambda}
    \left(\hat{\mathbf{x}}_i(t) - \mathbf{x}_i(t)\right)^\text{T}
    \mathbf{W}
    \left( \hat{\mathbf{x}}_i(t) - \mathbf{x}_i(t)\right) + \right. \\
    & \left. +\frac{1}{1-\lambda}
    \left(\hat{\mathbf{x}}_{i_c}(t) - {\mathbf{x}}_i(t)\right)^\text{T}
    \mathbf{W}
    \left(\hat{\mathbf{x}}_{i_c}(t) - {\mathbf{x}}_i(t)\right)
    \right)
\end{split}
\end{align}

The optimization has been solved in CasADi MATLAB \cite{andersson2019casadi} using the \textit{iptop} non-linear solver \cite{wachter2006implementation}. The parameters of the system are given in Table~\ref{table:param}, where the estimated ones are in the second part of the table.
% \begin{table}[!h]
% \begin{center}
% \begin{tabular}{c||c}
%     \textbf{parameter} & \textbf{value}  \\
%     \hline \hline
%      $m_{i}$& $1.16\cdot10^{-3} \text{kg}$\\
%      $\length{1}$ & $0.1950\, \text{m}$\\
%      $\length{2}$ & $0.1942\, \text{m}$\\
%      $\length{3}$ & $0.1827\, \text{m}$\\
%      $\length{4}$ & $0.1943\, \text{m}$\\
%      $\length{5}$ & $0.1977\, \text{m}$\\
%      \hline \\
%      $k_1$ & $11.312\, \text{N/m}$\\
%      $k_2$ & $5.411\, \text{N/m}$\\
%      $k_3$ & $15.519\, \text{N/m}$\\
%      $k_4$ & $7.008\, \text{N/m}$\\
%      $k_5$ & $14.477\, \text{N/m}$\\
%      $c$ & $0.002\, \text{N s/m}$\\
% \end{tabular}
% \caption{Cable parameter values.}
% \label{table:param}
% \end{center}
% \end{table}
\begin{table}[b]
\begin{center}
\begin{tabular}{c|c||c|c}
    \textbf{parameter} & \textbf{value} &\textbf{parameter}&\textbf{value}  \\
    \hline \hline
     $m_{i}$& $1.16\cdot10^{-3} \text{kg}$ &
     $\length{1}$ & $0.1950\, \text{m}$\\
     $\length{2}$ & $0.1942\, \text{m}$ &
     $\length{3}$ & $0.1827\, \text{m}$\\
     $\length{4}$ & $0.1943\, \text{m}$ &
     $\length{5}$ & $0.1977\, \text{m}$\\ \hline 
     $k_1$ & $11.312\, \text{N/m}$ &
     $k_2$ & $5.411\, \text{N/m}$\\
     $k_3$ & $15.519\, \text{N/m}$&
     $k_4$ & $7.008\, \text{N/m}$\\
     $k_5$ & $14.477\, \text{N/m}$&
     $c$ & $0.002\, \text{N s/m}$\\
\end{tabular}
\caption{Cable parameter values.}
\label{table:param}
\end{center}
\end{table}
The plots of the measured points over time and the corresponding simulated state are reported in Fig.~\ref{fig:est_plots}, and the evolution of the error between the two is in Fig.~\ref{fig:est_err}; the average error coordinates along the whole execution are always lower than $0.02\rm{m}$

We evaluated the model’s ability to generalize to diverse scenarios by assessing its performance in describing the cable dynamics in varying operational conditions. To do so, we let 2 Craziflie 2.1 UAVs manipulate the
cable\textemdash see some pictures from the experimental tests in Fig.~\ref{fig:exp_1}. The Crazyswarm software framework was used for data communication, trajectory control of the UAvs, and state estimation~\cite{crazyswarm}. An offboard computer transmitted the pre-computed reference positions (calculated offline starting from the desired positions of the outputs $\pos^d_3(t)$ and $\pos^d_4(t)$ using the proposed method) to the UAVs via radio communication. The external motion capture system provides position feedback to the UAVs.
We performed two types of tests:
\begin{itemize}
    \item Test A: rest-to-rest unidirectional flight. Being $\pos_{i,x}^d$ the $x-$ coordinate of the $\ith{i}$ desired position, the output desired trajectories for $i=3,4$ is
    $
    {\pos_{i,x}^d = x^i_{\text{0}}-x^i_{\text{a}}e^{-\frac{\left(t-t_{\text{0}}\right)^2}{C_\text{x}}},}$
 where $x^i_{\text{0}}$ is the initial position offset, $x^i_{\text{a}}=1.5\rm{m}$ the amplitude, $t_0$ the initial time offset, and $C_x$ the slope-tuning parameters.  Sub-tests have been performed with a distance between the outputs resulting in an inter-UAV distance of $0.3\rm{m}$ and $0.5\rm{m}$; in the following, the corresponding tests are referred to as \textit{narrow} and \textit{wide}, respectively. For each of them, two different velocities have been tested:  maximum UAV velocity is $1.70$ m/s$^{-1}$ in the \textit{slow} case ($C_x=1\rm{s}$) and $2.60$  m/s$^{-1}$ in the \textit{fast} case ($C_x=0.75\rm{s}$). Multiple consecutive exponential trajectories have been tried for each sub-test. 
 %\item Test B: 2D eight-shaped flight. In this test, the two output points are to perform an eight-shaped trajectory at a constant altitude where  
%$   { \pos_{i,x}^d= \pos_{i,y}^d\sin{(\omega_i t)}}$, 
 %${\pos_{i,y}^d=A_y\sin{(\omega_i t)}}$ with $A_y=0.75\rm{m}.$
 %Two sub-tests have been performed using a frequency of the sinusoidal functions equal to $\omega_i=0.6\rm{rad/s}$ and $\omega_i=0.9\rm{rad/s},$ obtaining a slower and faster execution, respectively.
 \item Test B: 3D eight-shaped flight. In this test, the two output points are to perform an eight-shaped trajectory 
$   { \pos_{i,x}^d= \pos_{i,y}^d\sin{(\omega_i t)}}$, 
 ${\pos_{i,y}^d=A_y\sin{(\omega_i t)}}$ with $A_y=0.75\rm{m}.$ $   { \pos_{i,z}^d= A_z\sin{(\omega_i t)}}$. Narrow ($A_z=0.010\rm{m}$) and wide ($A_z=0.015\rm{m}$) tests have been performed. $\omega_i=0.125\rm{rad/s}.$
\end{itemize}
The model performance is evaluated by feeding the measured UAV trajectories to the simulated model and comparing the output of the simulation ($\bm{p}_i^{sim}$) against the measured one ($\bm{p}_i^{meas}$). This allows accounting only for the error due to the model, disregarding the component due to the UAV tracking error.
%%%%%%%%%%%%%%%%%%%%
\begin{figure*}[t]
    \centering
\subfloat[\label{sf:closed_loop_a}]{\includegraphics[width=0.45\linewidth,height=9.5cm]{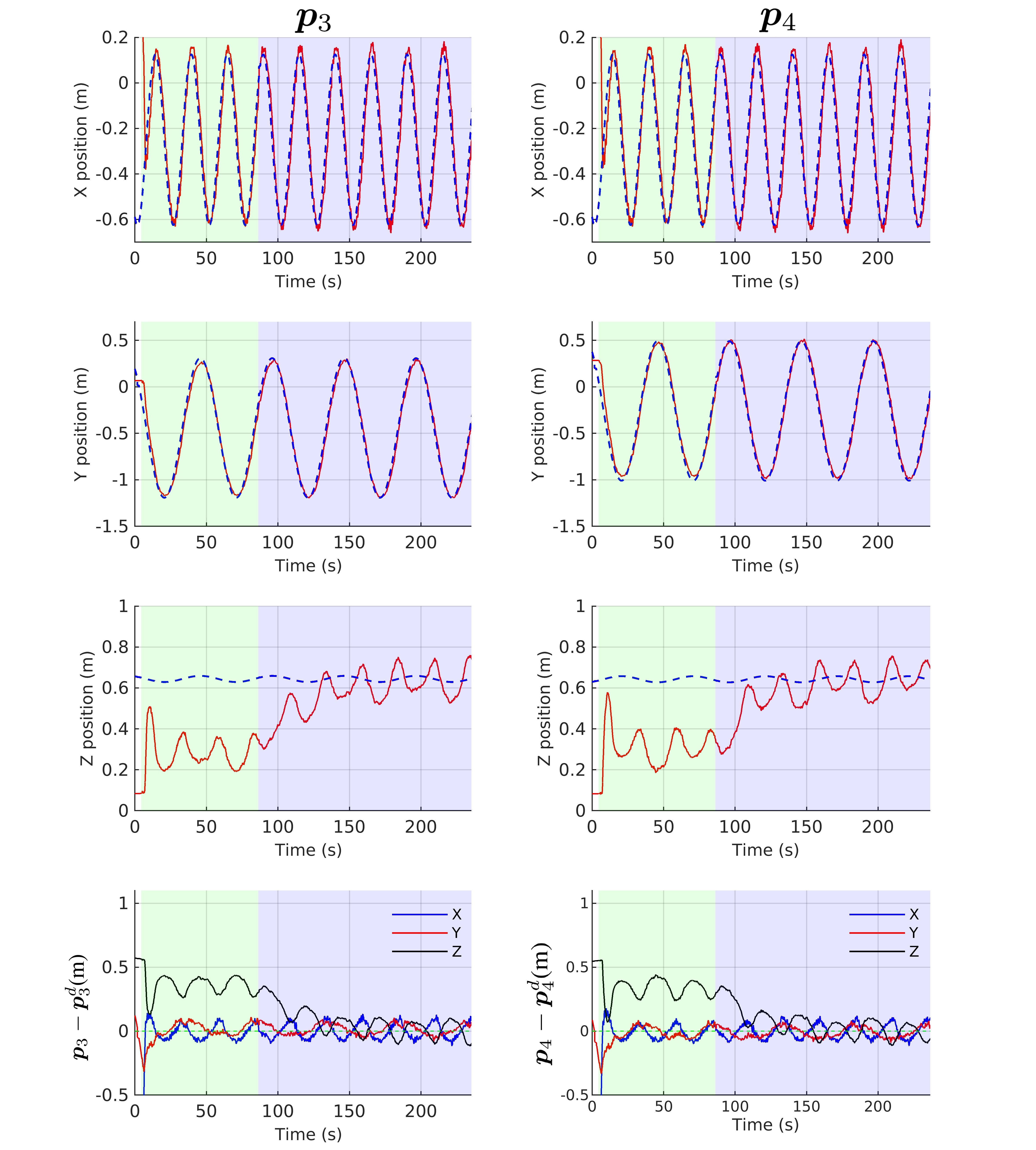}}\quad
\subfloat[\label{sf:closed_loop_b}]{\includegraphics[width=0.45\linewidth, height=9.5cm]{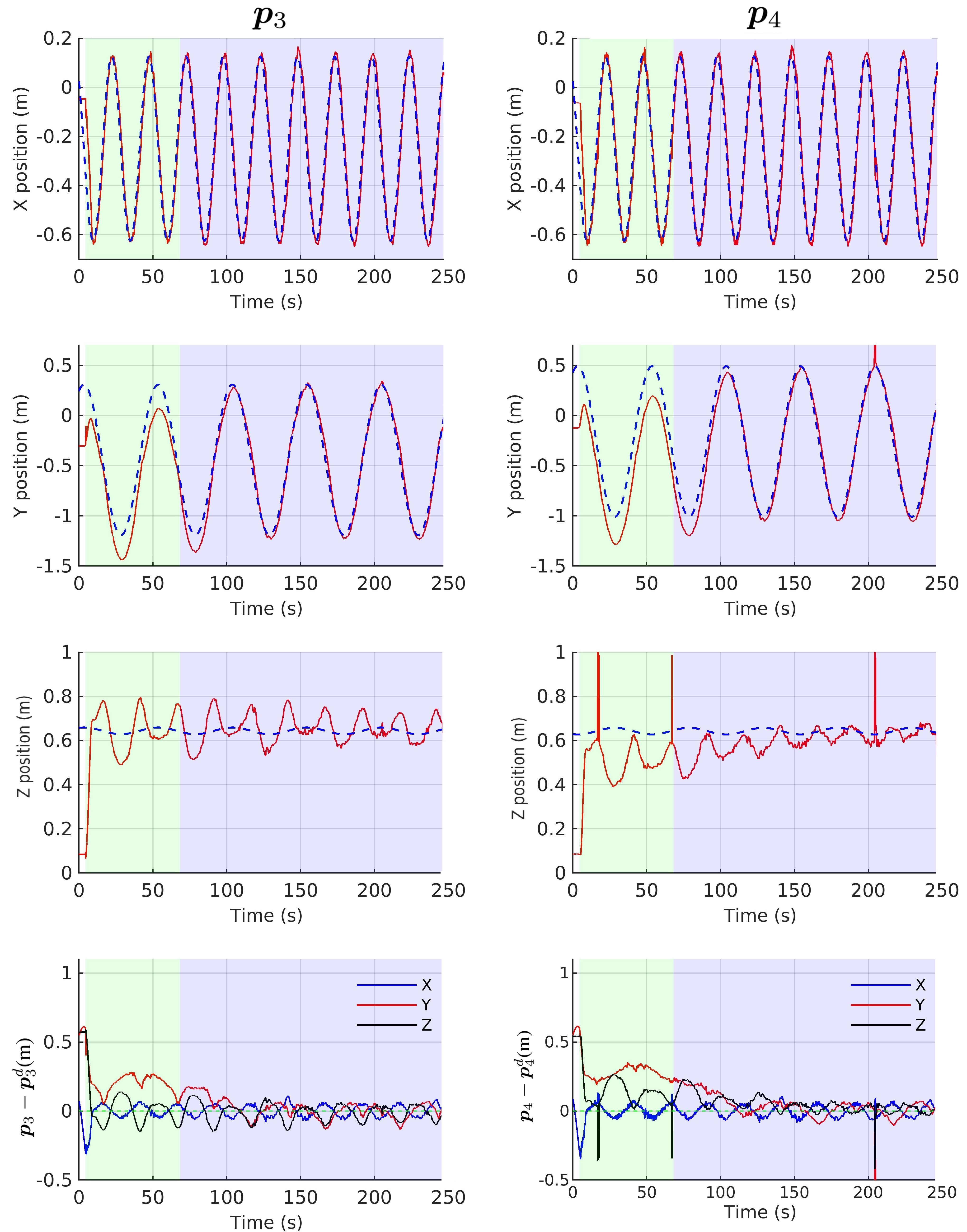}}
    \caption{Two (test B-narrow) closed-loop experiments where parameter uncertainties are induced by using a different cable. The control loop is closed using the output feedback in the blue region, where the output error is effectively reduced. Dashed lines are the reference values.  }
    \label{fig:closed-loop}
\end{figure*}
%%%%%%%%%%%%%%%%%
\begin{figure}[!h]
    \centering
\subfloat[Open loop\label{test2_open}]{\includegraphics[trim=0cm 0cm 0cm 2cm, clip,width=0.9\linewidth]{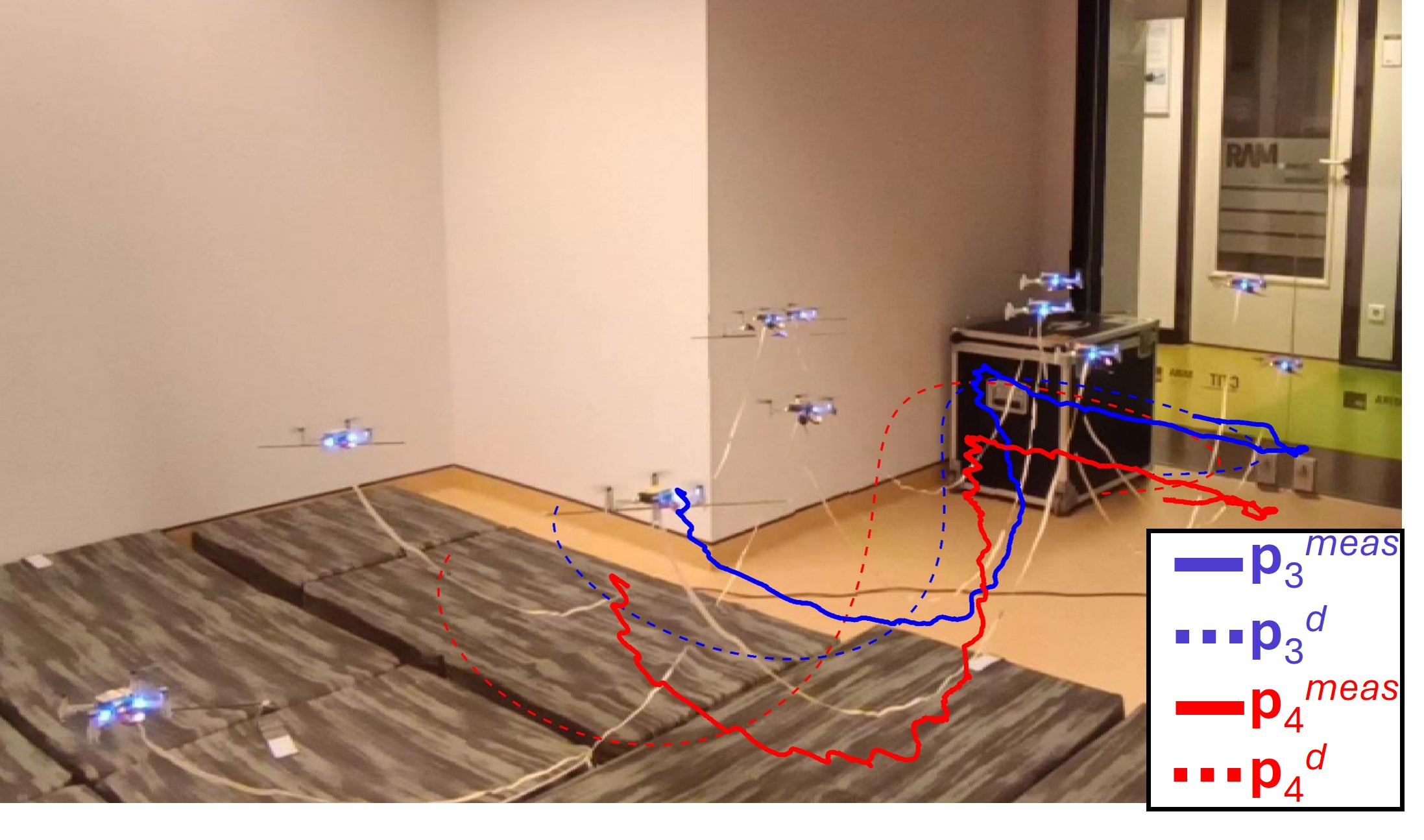}}\\
\subfloat[Closed loop\label{test2_closed}]{\includegraphics[trim=0cm 0cm 0cm 1cm, clip,width=0.9\linewidth]{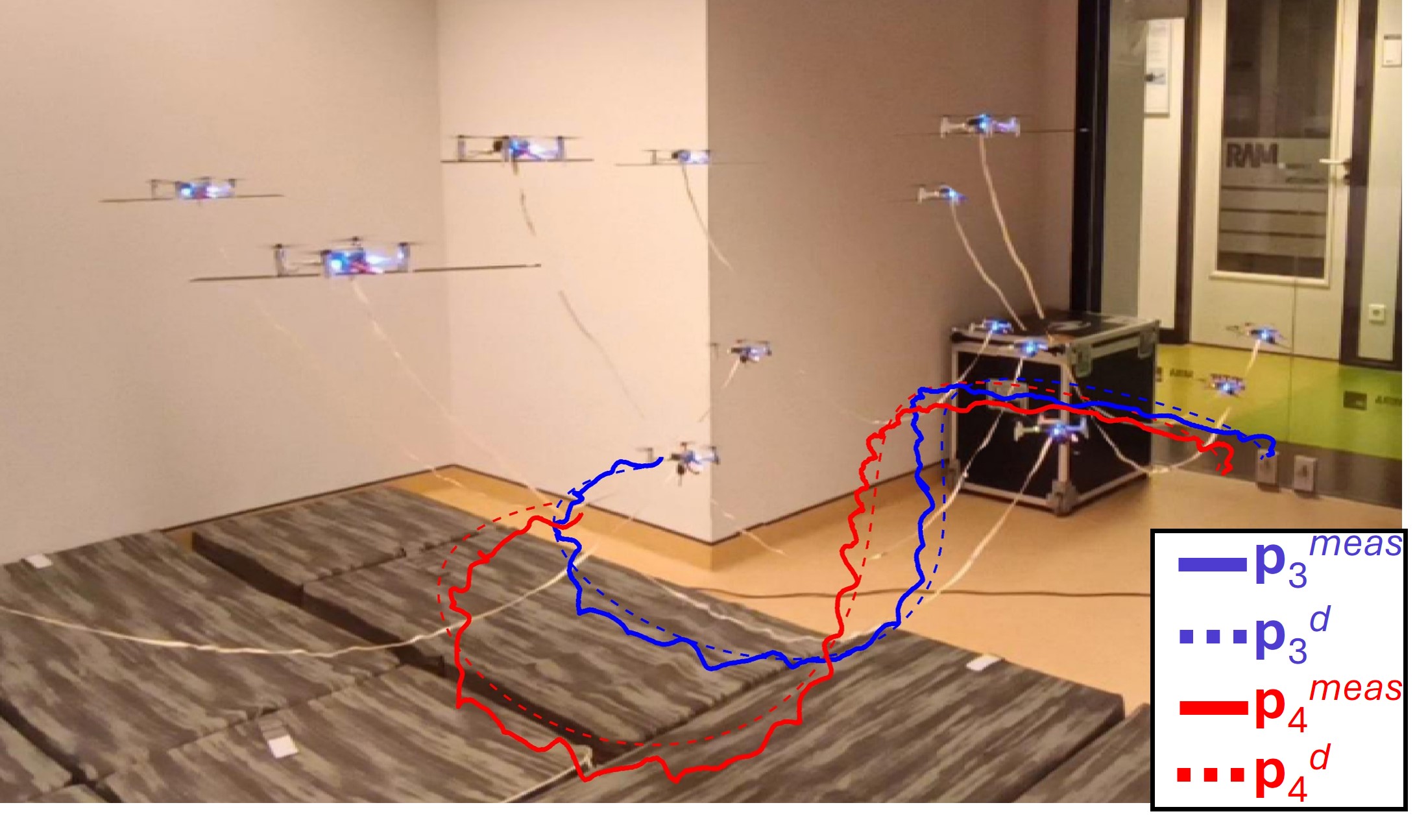}}
    \caption{Superimposed instants of part of the experiment in Fig.~\ref{sf:closed_loop_b}. Measured and desired output positions are highlighted and closer to each other with feedback control.}
    \label{fig:enter-label}
\end{figure}
For the sake of space, only the results of the narrow-fast sub-test A and the fast sub-test B are reported in Fig. \ref{fig:open_loop_tests}, but the average errors computed over all the sub-tests are reported in Table \ref{tabtraj1:errorTable}.  
\begin{table}[!h]\begin{center}
\begin{tabular}{|c||c|c|c|c|}
\hline
   \textbf{Test A} & \textbf{narrow slow} & \textbf{narrow fast} &\textbf{wide slow} &\textbf{wide fast} \\
    \hline \hline
    $\overline{{\mathbf{e}}_3}$ & 0.044 m	& 0.071 m &	0.037 m &	0.050 m \\
    \hline
    $\overline{\mathbf{e}_4}$ & 0.038 m & 0.057 m &	0.029 m &	0.041 m \\
    \hline
\end{tabular}\end{center}
\begin{center}
\begin{tabular}{|c||c|c|}
\hline
    \textbf{Test B}& \textbf{narrow} & \textbf{wide}   \\
    \hline \hline
    $\overline{\mathbf{e}_3}$ & 0.030 m	& 0.027 m \\
    \hline
      $\overline{\mathbf{e}_4}$ & 0.014 m	& 0.021 m \\
    \hline
\end{tabular}\end{center}
\label{tabtraj2:errorTable}
\caption{$\overline{\bm{e}_i}$ is the average of $||\bm{p}^{sim}_i-\bm{p}^{meas}_i||_2$, $i={3,4}$.}
\label{tabtraj1:errorTable}
\end{table}
% \begin{center}
% \begin{tabular}{|c||c|c|}
% \hline
%    \textbf{Test B} & \textbf{slow}  & \textbf{fast}  \\
%     \hline \hline
%     $||\bar{\mathbf{e}}_3||$ & 41.5 mm	& 54.4 mm  \\
%     \hline
%     $||\bar{\mathbf{e}}_4||$ & 37.8 mm	& 42.6 mm \\
%     \hline
% \end{tabular}
% \label{tabtraj2:errorTable}
% \end{center}
% \begin{center}
% \begin{tabular}{|c||c|c|}
% \hline
%     \textbf{Test C}& \textbf{narrow} & \textbf{wide} \\
%     \hline \hline
%     $||\bar{\mathbf{e}}_3||$ & 28.9 mm	& 27.4 mm  \\ $||\bar{\mathbf{e}}_4||$ & 14.5 mm	& 20.9 mm \\
%     \hline
% \end{tabular}
% \end{center}
% \caption{Average error between estimated and measured values for both output points in each test category.}
% \label{tabtraj1:errorTable}
% \end{center}
% \end{table}

\subsection{Cable Manipulation Control}\label{Sec:cable_manip_closed-loop}
In this section, we assess the performance of a cable-control strategy, based on the discrete cable model, closing the control loop on the cable outputs. We test a simple controller that integrates into the computation of the UAV trajectories, based on the differential flatness as explained in Section \ref{sec:flatness}, an integral term on the output error. That is similar to what we proposed in \cite{gabellieri2023differential} for systems $(c1)$ and tested solely through simulations so far. 
Specifically, given as the two output the positions of two consecutive points on the discrete cable model, $\pos_i$ and $\pos_{i+1}$, and given their desired trajectories as well, we compute the corresponding desired force $\cableForce{i}^d$ and we arrive at computing $\cableForceEq{i+1}^d$. We define $\err_i=\pos_{i}^d-\pos_{i}$ and write the analogous of \eqref{eq:f1_3} as ${\pos_{i+2}^d=\pos_{i+1}^d+\frac{\cableForceEq{i+1}^d}{\springCoeff{i+1}}-\length{i+1}\frac{\cableForceEq{i+1}^d}{||\cableForceEq{i+1}^d||} + K^I\int_0^t\err_{i+1}(\tau)\rm{d}\tau,}$ %\textcolor{red}{@YAOLEI, please, double-check that the controller is the one you used in the experiments; if not, modify the equation to make it equal to the one used in the experiments.}
where $\tau$ is an auxiliary variable, and $K^I\in\nR{3\times3}$ is a positive-definite diagonal matrix with elements on the diagonal equal to $0.2$ in the tests. The same procedure is applied when computing $\pos_{i-1}$ and so on, proceeding along the other side of the cable. Note that, with this method, the UAV references are continuously recomputed online based on output feedback. %\textcolor{red}{@YAOLEI: PLEASE, ADD DETAILS IF DIFFERENT FROM THE OPEN LOOP EXP. The controller runs offboard and the reference inputs are sent at XX Hz via radio communication.} 
The controller runs offboard and the references are sent at 100 Hz via radio communication to the quadrotors. The use of a different cable and different output point locations induces parameter uncertainties. In this first closed-loop test, a $3.6{\rm g}$ weight, $1.55{\rm m}$ long cable is iterated as the controlled object, while the two output points are located in the middle of the cable with $0.22{\rm m}$ distance. In the second closed-loop test, a $2.6{\rm g}$ weight, $1.24{\rm m}$ long cable is implemented, while the two output points are shifted toward one side of the cable.
%\textcolor{red}{mass .. kg and length .. m @YAOLEI PLEASE, PROVIDE VALUES}. 
The results of two closed-loop experiments are in Fig.~\ref{fig:closed-loop}, in which the same reference output as in Test B (narrow) was used.  The experimental results as well as the simulations can be found in the video attachment. 

\section{Discussion}
The proposed model showed an average error on each coordinate of the output lower than $0.02\rm{m}$ in the dataset used for parameter identification. 
During validation test A, the wide
configurations outperformed the narrow ones in both execution velocities. Intuitively, an increased stretch in the cable reduces its dangling motion.
The results also show that increasing the velocity of the execution increases the error.
 Overall, the model performance was satisfactory in the validation tests A and B, with the highest norm of the average output error registered in the narrow and fast test A and equal to $0.071\rm{m}$ for $\pos_3$ and $0.057\rm{m}$ for $\pos_4.$ Part of the error would be attributable to the sensor (the motion capture system in this case); however, that can be neglected here as its accuracy is a few millimeters.
Note that all available data were used for the parameter estimation with no division between training and validation datasets. This is justified by the low order of the discrete model, chosen \textit{a priori} for the sake of efficiency. However, high-order models, possibly requiring different data-driven estimation methods, would require careful handling of the training dataset to test against overfitting. 

The closed-loop experiments % \textcolor{red}{AF: what are CL experiments of type B? I don't see them defined before (only Test B in the identification section but to me that is a different test unless I missed something)}
showed that the controller is able to reduce the output error effectively. In the experiment shown in Fig.\ref{sf:closed_loop_a}, before the controller was activated, the average tracking error was  $0.373\rm{m}$ and  $0.347\rm{m}$ for $\pos_3$ and $\pos_4$, respectively. After the controller was switched on, the values decreased to $0.108\rm{m}$ and $0.100\rm{m}$. Similarly, for the experiment reported in Fig. \ref{sf:closed_loop_b}, the open-loop tracking errors for $\pos_3$ and $\pos_4$ were $0.216\rm{m}$ and $0.324\rm{m}$ and decreased to $0.088\rm{m}$ and $0.095\rm{m}$ when closing the control loop.
\section{Conclusions}\label{sec:conc}
The work demonstrated the differential flatness of a large class of aerial manipulation systems composed of single or multiple quadrotors connected to elastic and deformable cables. Simulation results supported the theoretical findings. Moreover, an experimental validation was carried out, demonstrating the suitability of the discrete model to describe a real system and to be used in a feedback control law.
In the future, the differential flatness will be exploited to design a more sophisticated model-based control, and a proof of stability will be drawn. Methods to optimize the number and distribution of cable segments based on the target task will be investigated. The applicability of the method to complex deformable objects will be studied, too.
\bibliographystyle{ieeetr}
\bibliography{Custombib}
\end{document}